\documentclass[11pt]{article}

\usepackage[final]{acl}

\usepackage{times}
\usepackage{latexsym}

\usepackage{microtype}
\usepackage{hyperref}
\usepackage{url}
\usepackage{booktabs}
\usepackage{amsfonts}
\usepackage{amsmath}
\usepackage{graphicx}
\usepackage{multirow}
\usepackage{fancyvrb}
\usepackage{xcolor}
\usepackage{tabularx}
\usepackage{float}
\usepackage{graphicx}
\usepackage{subcaption}

\usepackage{paralist}

\usepackage{cleveref}
\Crefname{equation}{Eq.}{Eqs.}
\Crefname{figure}{Fig.}{Figs.}
\Crefname{tabular}{Tab.}{Tabs.}
\Crefname{section}{\S}{\S}

\usepackage{hyperref}
\usepackage{url}
\usepackage[most]{tcolorbox}
\usepackage{booktabs}
\usepackage{graphicx}
\usepackage{lipsum}
\usepackage{amsmath}
\usepackage{subcaption}

\usepackage{fvextra}
\usepackage[most]{tcolorbox}

\usepackage[T1]{fontenc}

\usepackage[utf8]{inputenc}

\usepackage{microtype}

\usepackage{inconsolata}

\usepackage{graphicx}

\usepackage{xspace}

\newcommand{\ex}[1]{\textit{#1}\xspace}

\title{Failing to See or Failing to Know?\\ Attributing Errors in Vision-Language Models}

\author{
  \textbf{Khang Nhat Hoang Vo\textsuperscript{1}}
  \quad
  \textbf{Artem Vazhentsev\textsuperscript{1}}
  \quad
  \textbf{Artem Shelmanov\textsuperscript{1}}
\\
  \textbf{Timothy Baldwin\textsuperscript{1,2}}
  \quad
  \textbf{Yova Kementchedjhieva\textsuperscript{1}}
\\
\\
  \textsuperscript{1}MBZUAI
  \quad
  \textsuperscript{2}The University of Melbourne
\\
  \small{
    \textbf{Correspondence:}
    \href{mailto:Khang.Vo@mbzuai.ac.ae}{Khang.Vo@mbzuai.ac.ae}
    \quad
    \href{mailto:Yova.Kementchedjhieva@mbzuai.ac.ae}{Yova.Kementchedjhieva@mbzuai.ac.ae}
  }
}

\begin{document}
\maketitle

\begin{abstract}
Vision-language models (VLMs) can recognize entities in clear images yet still fail when answering questions that require factual knowledge beyond what is directly observable. Prior work has either examined individual failure modes in isolation or treated incorrect answers as monolithic, binary failures. We propose a tree-structured framework that organizes failures in knowledge-intensive visual question answering into model-specific operational outcomes. Across two datasets and four VLMs, we observe consistent
distributions of operational outcomes: some failures occur before entity recognition, while others persist after the relevant entity is recognized. Visual token representations are most informative
for recognition-related decisions. Prompt hidden states predict answer success more effectively, although factual-access attribution remains difficult and exhibits only a weak signal. These pre-generation signals support attribution-guided routing to targeted interventions, including image repair, entity support, question rewriting, and factual evidence.

\end{abstract}


\begin{figure}[t] 
\centering
\includegraphics[width=\linewidth]{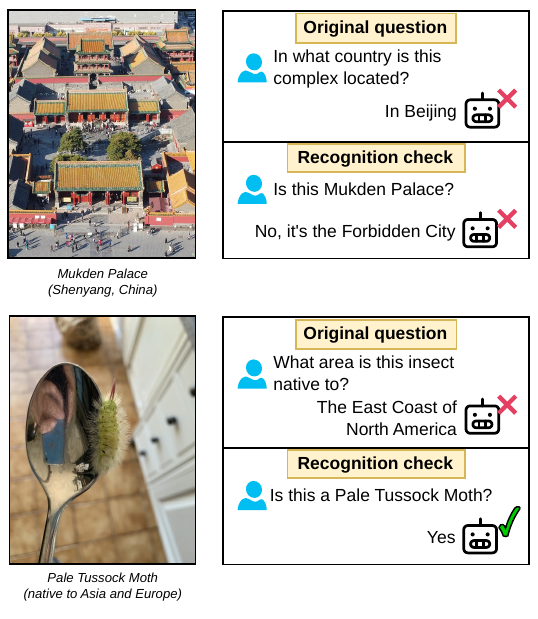}
\caption{Two recognition outcomes following an incorrect answer.
Top: the target VLM also fails the recognition check, yielding
\textsc{unknown entity}. Bottom: the target VLM recognizes the
entity, so attribution continues to the post-recognition factual
decisions.}
\label{fig:intro-three-cases} \end{figure}

\section{Introduction}

Modern vision-language models (VLMs) excel in visual perception over high-quality natural images \cite{li2023blip2, liu2023llava}.  However, users of assistive technologies, educational tools, search systems, and digital assistants often ask questions that go beyond what is directly visible, such as \ex{Where is this actor from?} or \ex{What is the natural habitat of this plant?} This form of \textit{knowledge-intensive} visual-question answering \citep{schwenk2022okvqa, marino2019ok, chen-etal-2023-pre-trained} requires both visual recognition and 
factual knowledge about the recognized entity. Failures can therefore arise from different stages 
of the image-to-answer process. \Cref{fig:intro-three-cases} illustrates two distinct paths to an incorrect answer. Different types of failures require different interventions. For example, if an image is of low quality, we should perform image repair or ask the user to submit a clearer one. If the question is too ambiguous, we should ask for a rewrite of the question. If a VLM cannot identify the object, we should use a more capable model. 

Prior work has either examined individual failure modes in isolation or treated incorrect answers as monolithic failures. Existing approaches include image-quality assessment and repair for degraded inputs~\citep{cai2025degradation}, evaluations of distributional gaps in entity recognition~\citep{liu2025culturevlm}, mechanistic analysis of failures to access known facts from visual entities~\citep{cohen-etal-2025-performance,venhoff2025toolate}, and retrieval augmentation when parametric knowledge is insufficient~\citep{lewis2020retrieval}. A complementary approach is abstention, which identifies and suppresses unreliable predictions regardless of the underlying issue, thus improving correctness at the expense of coverage~\citep{chandu2025certainlyuncertain}. Rather than making a binary decision about whether the model should answer, we seek to identify which local decision in the knowledge-intensive VQA process is associated with the failure. We test whether pre-generation representations can distinguish among failure types and support routing to targeted interventions before decoding even begins.

We turn failure attribution into a sequence of concrete diagnostic questions about each image–question pair: Does the target VLM recognize the depicted entity? If not, was recognition lost due to degraded visual evidence? If the entity is recognized, is the answer correct? If not, does the model succeed once the entity is named explicitly? Each question is answered by a controlled behavioral test of the target VLM, and together the tests form an attribution tree (\Cref{fig:attribution-tree}) that assigns every example a label identifying where it failed. These labels supervise a lightweight probe at each decision of the tree, trained on the VLM's internal representations using only the examples that reach that decision. Because these representations are available before the model generates an answer, the probes can predict the likely failure in advance and route the input to the matching intervention: image repair, entity-recognition support, question rewriting, or external factual evidence.



Our main contributions are as follows:
\begin{compactitem}
\item We introduce an attribution-tree framework for operational failure attribution in knowledge-intensive visual question answering, separating entity recognition, controlled visual-evidence failure, answer success, and factual access.
\item We show that likely failure outcomes can be predicted before any answer tokens are generated. Pre-generation visual-token features are strongest for recognition-related tasks, while prompt-boundary hidden states are more informative for post-recognition factual tasks.
\item We provide a proof-of-concept attribution-guided intervention experiment, showing that routing examples to the intervention matching the predicted failure type improves final answer accuracy by 31--39 percentage points.
\end{compactitem}

\section{Related Work}

Knowledge-intensive VQA \citep{schwenk2022okvqa, marino2019ok, chen-etal-2023-pre-trained} addresses information-seeking questions grounded in an observed image. The task is often seen as a two-hop process 
since it implicitly requires first identifying the relevant entities in the image and then retrieving the necessary facts about these entities \citep{venhoff2025toolate}. The multimodal nature of the task introduces various challenges within and across modalities, which are briefly outlined below.

\subsection{Failure Modes and Mitigation}

\paragraph{Low-quality images.}
\citet{cheng2025understanding} find that VLMs are rarely able to flag artificially-degraded medical images as having poor quality, even when prompted to do so. Instead, VLMs proceed to make an ungrounded and inaccurate diagnostic prediction. Similarly, for chart-based reasoning, \citet{shin2025losing} find that VLMs make overconfident wrong predictions in degraded settings. To mitigate this, \citet{cai2025degradation} train a classifier to label images with distinct degradation types (resolution issues, motion blur, etc.) and apply tailored image enhancement techniques. 

\paragraph{Unknown entities.} Recent work has studied the issue of unknown visual entities in terms of cultural diversity and underrepresentation. Over 16 VLMs, \citet{liu2025culturevlm} find that the models perform well in classifying and describing Western concepts but do poorly in describing concepts from Asia and Africa. \citet{nayak2024benchmarking} and \citet{romero2024cvqa} report similar findings, which amount to reduced entity recognition in out-of-distribution contexts: entities that are rare or absent in the pretraining data are harder for VLMs to recognize at inference time.

\paragraph{Cross-modal access.} It has been observed that VLMs tend to underperform on visually-grounded entity-centric questions compared to equivalent one-hop questions that state the entity explicitly, e.g.\ \ex{Where is this actor from?} vs.\ \ex{Where is Tom Cruise from?} \citet{venhoff2025toolate} and \citet{cohen-etal-2025-performance} study this via distinct forms of mechanistic interpretability, reaching the same conclusion: entity recognition happens too late in the decoder, blocking factual recall even for well-known facts. \citet{venhoff2025toolate} show that prompting the VLM to explicitly name the entity first and then answer the question improves performance, which serves both as evidence for the mechanism they uncover and as a viable mitigation strategy.   

\paragraph{Missing facts.} Lastly, a major obstacle to successful knowledge-intensive VQA can be the mere absence of a relevant fact from the parametric knowledge of the VLM. Retrieval augmentation \citep{lewis2020retrieval} can mitigate this effect \citep{yasunaga2022retrieval, lin-byrne-2022-retrieval, chen-etal-2023-pre-trained, qi2024rora, sravanthi-etal-2025-rg}. In much of this work, retrieval augmentation is applied indiscriminately, without checking whether it is needed \citep{asai2024selfrag, jiang-etal-2023-active}. A common approach to retrieving relevant context in entity-centric VQA is to first identify the relevant entity via an external module \citep{chen-etal-2023-pre-trained} or by captioning the image \citep{sravanthi-etal-2025-rg}, followed by text-to-text retrieval. 

Each of these issues is well documented and admits model- or tool-based mitigations. Yet they have largely been studied in isolation: in knowledge-intensive VQA, few methods distinguish among these error types and route each example to the appropriate intervention.


\subsection{Hallucination and Uncertainty}

Knowledge-intensive VQA failures overlap closely with multimodal hallucination, where a model generates content that is unsupported by the image or by factual knowledge~\citep{li2023hallucination,Guan_2024_CVPR,liu2024surveyhallucinationlargevisionlanguage}. Most prior work evaluates hallucination at the level of the generated response. POPE focuses on object mentions that are unsupported by the image~\citep{li2023hallucination}, while HallusionBench studies failures arising from the interaction between visual illusion and language-based reasoning~\citep{Guan_2024_CVPR}. Other benchmarks broaden the evaluation beyond object existence. M-HalDetect annotates hallucinated objects, attributes, and relations in detailed descriptions~\citep{10.1609/aaai.v38i16.29771}; AMBER covers both generative and discriminative hallucination settings without relying on LLM judges~\citep{wang2024amberllmfreemultidimensionalbenchmark}; and FaithScore decomposes responses into atomic image-grounded facts for fine-grained faithfulness evaluation~\citep{jing-etal-2024-faithscore}. MMHal-Bench evaluates unsupported content in multimodal instruction-following responses~\citep{sun-etal-2024-aligning}, while subsequent work develops unified benchmarks and tool-based detectors spanning multiple hallucination categories~\citep{chen-etal-2024-unified-hallucination}. MHALO further evaluates MLLMs as token-level hallucination detectors, distinguishing perception- and reasoning-related errors~\citep{cai-etal-2025-mhalo}.

Complementary work studies hallucination through visual-attention dynamics. \citet{Xie-ilvad-icml2026} use inter-layer attention discrepancies to recover relevant evidence during decoding, whereas we use pre-generation states to attribute likely failures before decoding. Closest to our setting, \citet{kogilathota-etal-2026-halp} predict hallucination risk from visual-token and query-token representations. Our goal instead is to identify which local decision is associated with the failure, including entity recognition, visual evidence, answer success, and factual access after recognition.

A second closely related direction studies uncertainty and reliability signals. In LLMs, \citet{sriramanan2024llmcheck} analyze hallucination detection using hidden states, attention maps, and output probabilities, while \citet{shelmanov-etal-2025-head} train auxiliary uncertainty-quantification heads for detecting hallucinated outputs. In VLMs, \citet{khan2024consistency} use neighborhood consistency to identify unreliable black-box responses, and \citet{chandu2025certainlyuncertain} introduce a benchmark and taxonomy for multimodal epistemic and aleatoric uncertainty, evaluating whether models recognize answerability under different sources of uncertainty.

Our work is related in its use of model-internal signals but differs in the prediction target. Rather than estimating a single hallucination or reliability score, we predict the local decisions in an operational attribution tree, separating entity recognition, visual evidence, answer success, and factual access.

\section{Methodology}

\subsection{Operational Failure Attribution}
\label{sec:preliminaries}

Our approach attributes knowledge-intensive VQA failures through a sequence of local diagnostic decisions rather than inferring a failure type from the final answer alone. For each image-question pair, controlled behavioral tests determine whether the target VLM recognizes the relevant entity, whether the image provides sufficient evidence for recognition, whether the original question is answered correctly, and, if not, whether the required fact becomes accessible when the entity is explicitly named. These decisions form a tree with five leaves: \textsc{visual-evidence failure}, \textsc{unknown entity}, \textsc{unknown fact}, \textsc{unrecallable fact}, and \textsc{success}.

We train one lightweight probe for each decision using representations extracted from the frozen target VLM, and compose the resulting probabilities along the tree to obtain scores over the five outcomes. These outcomes are operational, model-specific diagnostics derived from controlled tests; they should not be interpreted as causal ground truth or as evidence that every error has a single, uniquely identifiable source.

\begin{figure}[t]
    \centering
    \includegraphics[width=\linewidth]{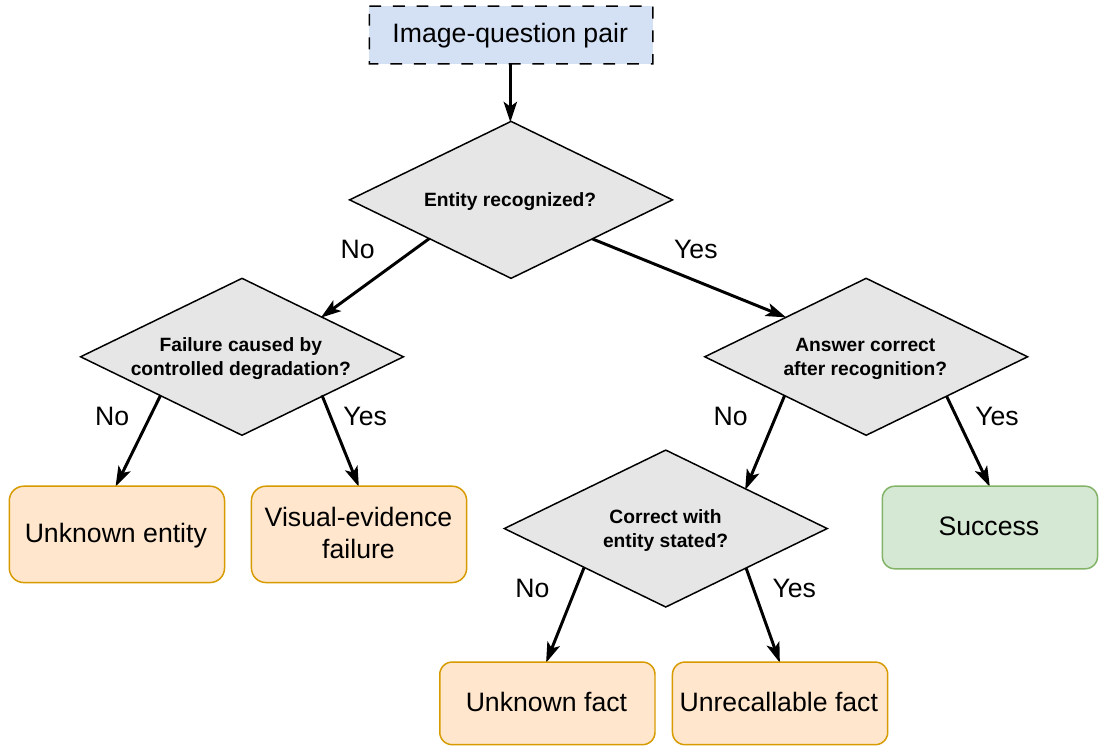}
\caption{Operational attribution tree for knowledge-intensive VQA.}
    \label{fig:attribution-tree}
\end{figure}

As shown in \Cref{fig:attribution-tree}, the five outcomes are composed from four local binary decisions:
\begin{compactitem}
    \item \textbf{Recognition:} whether the target VLM reliably recognizes the relevant visual entity.
    \item \textbf{Visual evidence:} among recognition failures, whether recognition was lost under controlled image degradation.
    \item \textbf{Answer success:} among recognized examples, whether the original factual question is answered correctly.
    \item \textbf{Factual access:} among recognized but incorrectly answered examples, whether the model succeeds once the entity is stated explicitly.
\end{compactitem}

This tree-structured formulation allows each local decision to use the representation most appropriate to it while retaining scores over the five attribution outcomes. It can also be interpreted from a reliability perspective: binary failure detection is recovered by summing the four non-success leaf probabilities. Our primary goal, however, is fine-grained failure attribution rather than estimating a single reliability score.

\subsection{Data Curation}
\label{sec:data_curation}

We sample image-question pairs from two existing knowledge-intensive VQA datasets: PopVQA \citep{cohen-etal-2025-performance}, consisting of factual image-question pairs across four types of popular entities, including celebrities, landmarks, logos, and paintings; and the iNaturalist subset of Encyclopedic VQA \citep{vanhorn2018inaturalist, mensink23iccv}, a dataset of factual questions about plant and animal species. We sample 6,300 questions from PopVQA and 4,400 questions from iNaturalist.

To focus on image-conditioned behavior and complex factual recall, we apply two filters. First, we exclude binary questions (starting with \emph{Does}, \emph{Can}, or \emph{Is}) and disjunctive questions containing explicit \emph{or}. 
Second, we filter for visual grounding using an image-ablation test: we replace the image with a white canvas and ask the same factual question. We remove an item if \textit{any} target VLM answers correctly under
image ablation for any evaluated answer permutation. 
After filtering, the dataset contains $4{,}863$ PopVQA pairs and $2{,}116$ iNaturalist VQA pairs. Some sample entity-linked questions can be seen in Appendix \ref{app:sample-qa}.


Below, we describe how we identify instances of the different failure modes in the predictions of a single target VLM given a VQA pair.


\paragraph{Unknown entity.}
Some examples may contain entities that the target VLM does not reliably recognize under our verification protocol: less well-known celebrities, rare plants, etc. To establish whether the VLM reliably recognizes the main entity in the input image, we run a simple verbalized recognition probe. In early experiments, we found it challenging to evaluate responses to an open question such as ``\ex{What is the plant in the image?}''. Given the importance of correctly classifying failure modes for our purposes, we instead opted for a more controlled \textit{yes/no} setting, where we asked ``\ex{Is the plant in the image [plant name]?}''. To reduce sensitivity to chance performance and affirmative response bias, we construct three distractor prompts per ground-truth prompt, replacing the true entity with a randomly sampled entity from the same subtype (e.g., one mammal replaces another mammal). If the VLM answers \textit{yes} to the ground-truth prompt and \textit{no} to all three distractors, we consider the entity to be reliably recognized by the model. Otherwise, the example receives the label \textsc{unknown entity}.

\paragraph{Visual-evidence failure.}
The datasets we build on (PopVQA and iNaturalist) do not contain instances of low-quality images by design; yet, in the wild, this can be a common problem and thus should be accounted for. We construct a controlled visual-evidence stress test by
progressively applying Gaussian blur, additive Gaussian noise,
JPEG compression, and down-sampling, followed by resizing back to the model input size -- with severity gradually increasing (further details in Appendix \ref{app:visual-evidence-construction}). After each corruption step, we reprompt the VLM to recognize the ground-truth entity. This process produces two possible outcomes. In some cases, the model continues to recognize the entity even under the maximum corruption level; these examples remain in the recognizable branch and are not labeled as visual-evidence failures. In other cases, the model changes from recognizing the entity to failing to recognize it. We keep the first corrupted image at which this flip occurs and label it as a visual-evidence failure. Thus, visual-evidence failures are not arbitrary noisy images; they are originally recognizable examples whose recognizability is lost under controlled visual degradation. We then balance this class against the unknown-entity class defined above. Specifically, we downsample the corrupted visual-evidence failures so that their count matches the number of unknown-entity examples, yielding a balanced recognition-failure split. 
Accordingly, this branch should be interpreted as a controlled proof-of-concept stress test rather than a model of the full range of naturally occurring visual failures.

\paragraph{Unrecallable fact vs.\ unknown fact.}
Shifting focus to the second stage of the QA process, we isolate cases where, given successful entity recognition, the final answer is still wrong: we distinguish factual failures that are resolved by an entity-explicit question from those that persist under this intervention. Following the methodology of \citet{cohen-etal-2025-performance}, we identify failures in cross-modal recall by testing VLMs on a version of the question that explicitly mentions the entity name, e.g., ``\ex{What year was Tom Cruise born?}'' instead of ``\ex{What year was this actor born?}'' (with the image still available, although the model no longer needs to infer the entity name from it). If the model fails on the original question but succeeds on the
entity-explicit version, we assign the
\textsc{unrecallable fact} label, operationally indicating that the failure is resolved when the entity is stated explicitly. If the model fails on both versions, we assign the
\textsc{unknown fact} label, indicating that the failure persists under
the entity-explicit intervention. To assess response correctness, we use an LLM judge and provide both the VLM response and the gold-standard answer provided in the datasets for reference (see Appendix~\ref{app:eval_judge} for more details).

Once every VQA pair from PopVQA and iNaturalist is processed by a given VLM and assigned to one of the failure modes above, we obtain a model-specific dataset for training probes to predict probabilities over the different operational outcomes.

\subsection{Features for Probes}

We extract pre-generation features after the image--question prompt has been processed and before answer generation begins. \textsc{Vis} denotes decoder-side visual-prompt features taken from the hidden state at the final image-token position, motivated by prior work showing that VLM image-token representations encode localized visual information and support object and attribute grounding~\citep{11094641}. \textsc{EOP} denotes the hidden state of the final prompt token, commonly used as a compact representation for probing model behavior and reliability~\citep{zhang-etal-2025-prompt}. \textsc{Last8} concatenates the hidden states of the final eight prompt tokens, providing a small prompt-boundary context rather than a single-token representation. \textsc{Attn} consists of flattened attention weights from these eight tokens to a four-token lookback window. Such features summarize local prompt-context dependencies and are motivated by prior attention-based hallucination and
uncertainty detectors~\citep{chuang2024lookback,vazhentsev-etal-2025-unconditional}. \textsc{Attn+Last8} combines prompt-boundary hidden states with
local attention patterns to test whether attribution benefits from both state- and dependency-based signals. All features concatenate mid-to-late-layer offsets $(-1,-4,-8,-12)$, where $-N$ denotes the $N$th layer counted backward from the final decoder layer. Relative offsets keep feature definitions comparable across models with different decoder depths. Appendix~\ref{app:rq2-formal} provides formal definitions.

\section{Experiments}
\label{sec:experiments}

\subsection{Experimental Setup}

\paragraph{Choice of VLMs.}
We evaluate four widely used open-weight VLMs: Gemma-3-12B-it with 48 decoder layers \citep{gemma3}, Llama-3.2-11B-Vision-Instruct with 40 layers \citep{llama32vision}, Qwen2.5-VL-7B-Instruct with 28 layers \citep{qwen25vl}, and Qwen3-VL-8B-Instruct with 36 layers \citep{qwen3vl}. They cover diverse model families and multimodal designs while remaining practical for reproducible inference and feature extraction.

\paragraph{Probe architecture.}
We experiment with two probe architectures: a lightweight linear classification head on top of the concatenated feature vectors, and a two-layer Transformer encoder applied to sequence-structured features.

\begin{figure}[t]
    \centering
    \includegraphics[width=\linewidth]{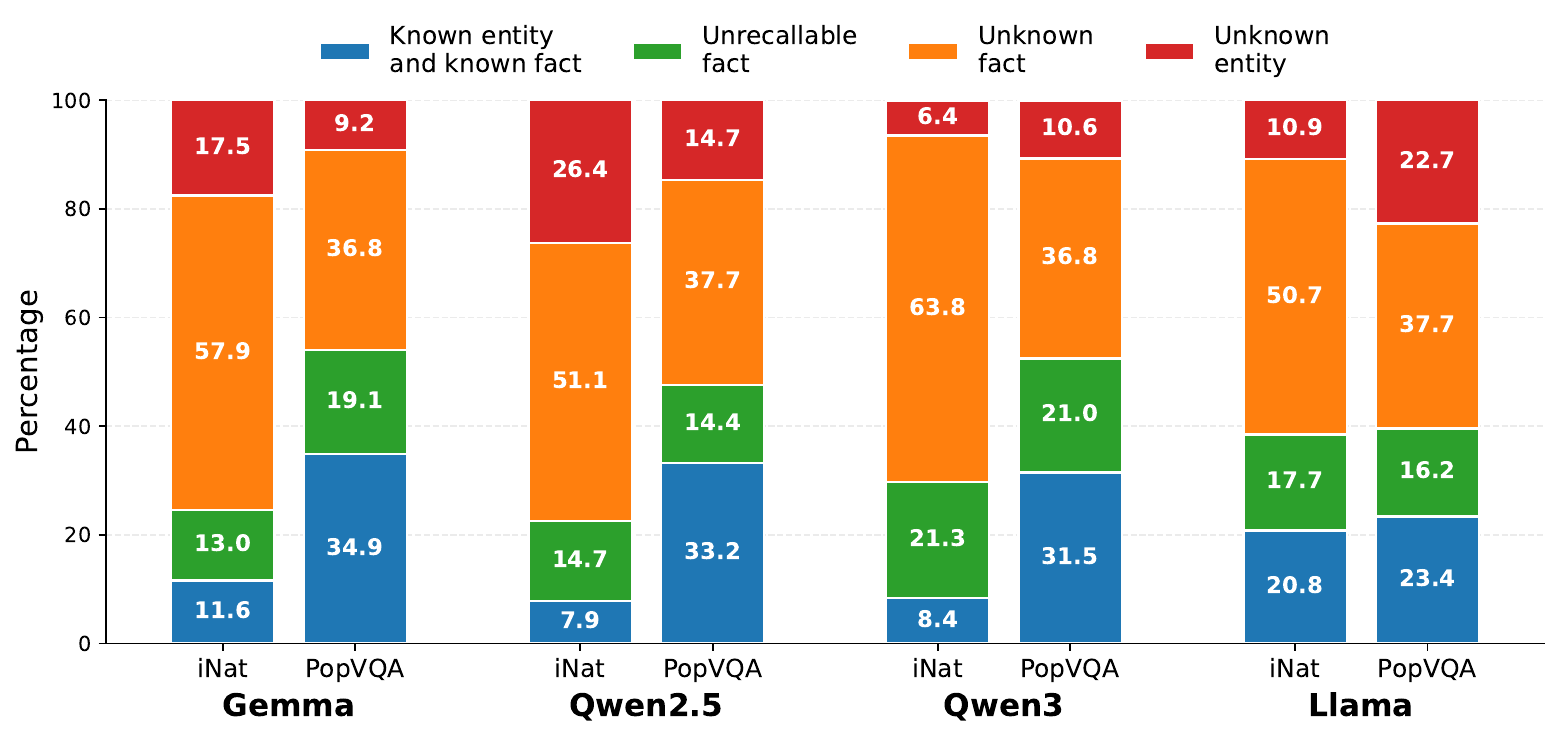}
    \caption{Distribution of attribution outcomes across models and datasets. 
    }
    \label{fig:rq1-4way}
\end{figure}

\paragraph{Training and Evaluation.}
We train four local binary probes following the attribution tree in \S\ref{sec:preliminaries}, with each probe corresponding to one decision point and trained only on examples that pass the preceding checks. Each decision-specific dataset is split into train, validation, and test sets with stratified sampling. Hyperparameters are selected on the validation split (detailed in Appendix \ref{app:grid-search}), and we report test PR-AUC, which accounts for class imbalance in the tree. All experiments are run on a single 40GB VRAM GPU.

\paragraph{Baselines.}
We compare our approach to several post-generation uncertainty quantification (UQ) baselines that operate after answer generation is completed and test whether the generated answer, decoding trace, or answer-level uncertainty exposes the  failure signal. \textsc{OutSeq} extracts attention patterns and top-$m$ next-token log-probabilities across the generated answer sequence, following prior work on hallucination detection from decoding-time representations~\citep{shelmanov-etal-2025-head}. \textsc{TokProb} uses decoding-time token probabilities as a lightweight confidence signal. 
We also evaluate standard UQ baselines for free-form generation, including \textsc{MSP}~\citep{hendrycks2017baseline}, \textsc{MTE} and \textsc{LexSim}~\citep{fomicheva-etal-2020-unsupervised}, \textsc{SemanticEntropy}~\citep{kuhn2023semantic}, \textsc{SAR}~\citep{duan-etal-2024-shifting}, \textsc{SemanticDensity}~\citep{qiu2024semanticdensity}, and \textsc{CoCoA} variants~\citep{vashurin2025cocoa}. 
We report \textsc{MSP} as a standard confidence baseline and \textsc{Oracle-UQ} as the best score among all evaluated UQ methods for each model and attribution decision. \textsc{Oracle-UQ} is an upper bound over answer-level UQ baselines rather than a deployable prediction strategy.

\subsection{Results}


\paragraph{Failure attribution analysis.}

We run each model on our curated PopVQA and iNaturalist pairs and assign attribution outcomes using the procedure in \S\ref{sec:data_curation}. As shown in \Cref{fig:rq1-4way}, all models have low success rates on both subsets: the best success rate is 20.8\% on iNaturalist for Llama and 34.9\% on PopVQA for Gemma.\footnote{We exclude the reduced visual evidence category because it is artificially constructed from recognized entities.} The dominant failure mode is usually \textsc{unknown fact}, accounting for 50.7--63.8\% of iNaturalist examples and 36.8--37.7\% of PopVQA examples, showing that many errors occur even after the entity is recognized. \textsc{Unknown entity} failures are also substantial but more model-dependent: they range from 6.4--26.4\% on iNaturalist and 9.2--22.7\% on PopVQA. Finally, \textsc{unrecallable fact} remains non-negligible, ranging from 13.0--21.3\% on iNaturalist and 14.4--21.0\% on PopVQA. Overall, knowledge-intensive VQA failures are not monolithic: some arise before entity recognition, while many remain on the post-recognition branch.

\paragraph{Pre-generation feature performance.}
\label{sec:53}

\begin{table*}[t]
\centering
\small
\renewcommand{\arraystretch}{0.82}
\resizebox{\textwidth}{!}{%
\begin{tabular}{llccccc|ccccc}
\toprule
& &
\multicolumn{5}{c|}{\textbf{Recognition:} recognized vs.\ unrecognized}
&
\multicolumn{5}{c}{\textbf{Visual evidence:} unknown entity vs.\ visual-evidence failure}
\\
\cmidrule(lr){3-7}
\cmidrule(lr){8-12}

\textbf{Method} & \textbf{Head}
& \textbf{Gemma}
& \textbf{Qwen2.5}
& \textbf{Qwen3}
& \textbf{Llama}
& \textbf{Avg.}
& \textbf{Gemma}
& \textbf{Qwen2.5}
& \textbf{Qwen3}
& \textbf{Llama}
& \textbf{Avg.}
\\
\midrule

\multicolumn{12}{l}{\textit{Pre-generation probes}} \\

\textsc{Vis} & Lin.
& \textbf{87.8}
& \textbf{91.3}
& 81.4
& \textbf{86.1}
& \textbf{86.7}
& \textbf{98.4}
& \textbf{99.7}
& \textbf{90.7}
& \textbf{99.3}
& \textbf{97.0}
\\

\textsc{EOP} & Lin.
& 82.3
& 72.0
& 85.3
& 72.4
& 78.0
& 62.0
& 60.3
& 61.6
& 62.5
& 61.6
\\

\textsc{Last8} & Lin.
& 83.8
& 75.7
& 85.1
& \underline{73.2}
& 79.4
& 67.2
& 65.5
& 59.9
& 71.1
& 65.9
\\

\textsc{Last8} & Tr.
& 84.2
& 76.3
& 86.1
& 71.1
& 79.4
& 66.6
& 64.9
& 61.2
& 70.1
& 65.7
\\

\textsc{Attn} & Lin.
& \underline{85.7}
& 74.3
& 85.2
& 72.5
& 79.4
& 66.8
& 65.5
& \underline{65.8}
& 68.4
& 66.6
\\

\textsc{Attn+Last8} & Lin.
& 83.9
& \underline{77.4}
& \textbf{87.1}
& 72.0
& \underline{80.1}
& \underline{68.4}
& \underline{67.0}
& 62.6
& \underline{72.3}
& \underline{67.6}
\\

\midrule
\multicolumn{12}{l}{\textit{Post-generation and uncertainty baselines}} \\

\textsc{OutSeq} & Lin.
& 80.5
& 75.5
& \underline{86.6}
& 65.0
& 76.9
& 56.2
& 59.4
& 54.2
& 53.7
& 55.9
\\

\textsc{OutSeq} & Tr.
& 83.1
& 73.6
& 84.9
& 67.8
& 77.3
& 57.5
& 60.7
& 55.7
& 55.1
& 57.2
\\

\textsc{TokProb} & Lin.
& 81.3
& 74.8
& 82.4
& 69.1
& 76.9
& 60.2
& 62.2
& 57.1
& 56.7
& 59.1
\\

\textsc{MSP} & --
& 80.6
& 72.6
& 85.1
& 70.4
& 77.2
& 37.3
& 36.4
& 30.0
& 29.0
& 33.2
\\

\textsc{Oracle-UQ} & --
& 81.2
& 73.8
& 85.1
& 70.5
& 77.7
& 37.3
& 36.8
& 30.4
& 31.6
& 34.0
\\

\textsc{Random} & --
& 78.5
& 71.0
& 82.1
& 66.3
& 74.5
& 32.6
& 34.1
& 30.2
& 28.7
& 31.4
\\

\bottomrule
\end{tabular}%
}
\caption{
Test PR-AUC for image-side attribution decisions, multiplied by 100.
Bold and underlined values indicate the best and second-best scores, respectively, in each column.
}
\label{tab:image_side_probe_results}
\end{table*}

\begin{table*}[t]
\centering
\small
\renewcommand{\arraystretch}{0.82}
\resizebox{\textwidth}{!}{%
\begin{tabular}{llccccc|ccccc}
\toprule
& &
\multicolumn{5}{c|}{\textbf{Answer success:} success vs.\ failure after recognition}
&
\multicolumn{5}{c}{\textbf{Factual access:} unknown fact vs.\ unrecallable fact}
\\
\cmidrule(lr){3-7}
\cmidrule(lr){8-12}

\textbf{Method} & \textbf{Head}
& \textbf{Gemma}
& \textbf{Qwen2.5}
& \textbf{Qwen3}
& \textbf{Llama}
& \textbf{Avg.}
& \textbf{Gemma}
& \textbf{Qwen2.5}
& \textbf{Qwen3}
& \textbf{Llama}
& \textbf{Avg.}
\\
\midrule

\multicolumn{12}{l}{\textit{Pre-generation probes}} \\

\textsc{Vis} & Lin.
& 36.1
& 37.4
& 34.9
& 35.7
& 36.0
& 33.9
& 32.6
& 31.2
& 35.3
& 33.2
\\

\textsc{EOP} & Lin.
& 52.9
& 54.1
& 57.8
& 51.7
& 54.1
& 47.3
& 45.2
& 50.3
& 46.9
& 47.4
\\

\textsc{Last8} & Lin.
& \textbf{61.2}
& \textbf{62.5}
& \textbf{66.8}
& \textbf{60.4}
& \textbf{62.7}
& 52.1
& 49.8
& \underline{55.7}
& 51.4
& 52.3
\\

\textsc{Last8} & Tr.
& 55.6
& 56.9
& 54.3
& \underline{53.2}
& 55.0
& 53.4
& \underline{50.8}
& \textbf{57.2}
& \underline{52.7}
& \textbf{53.5}
\\

\textsc{Attn} & Lin.
& 49.8
& 50.4
& 53.5
& 48.2
& 50.5
& 43.9
& 44.6
& 48.2
& 42.9
& 44.9
\\

\textsc{Attn+Last8} & Lin.
& 54.4
& 55.2
& \underline{58.7}
& 53.0
& 55.3
& 48.6
& 48.0
& 52.4
& 47.2
& 49.0
\\

\midrule
\multicolumn{12}{l}{\textit{Post-generation and uncertainty baselines}} \\

\textsc{OutSeq} & Lin.
& 52.6
& 53.5
& 55.4
& 50.1
& 52.9
& 49.6
& 47.8
& 52.0
& 48.3
& 49.4
\\

\textsc{OutSeq} & Tr.
& \underline{56.0}
& \underline{57.6}
& 58.4
& 52.8
& \underline{56.2}
& 51.8
& 49.9
& 54.4
& 50.6
& 51.7
\\

\textsc{TokProb} & Lin.
& 53.2
& 54.0
& 56.0
& 51.2
& 53.6
& 50.5
& 48.7
& 53.0
& 49.2
& 50.4
\\

\textsc{MSP} & --
& 30.5
& 28.3
& 34.3
& 26.9
& 30.0
& \underline{55.0}
& 48.8
& 47.0
& 41.0
& 47.9
\\

\textsc{Oracle-UQ} & --
& 30.7
& 30.1
& 35.4
& 29.5
& 31.4
& \textbf{55.9}
& \textbf{51.5}
& 48.7
& \textbf{53.5}
& \underline{52.4}
\\

\textsc{Random} & --
& 29.3
& 26.5
& 33.5
& 29.2
& 29.6
& 49.7
& 49.8
& 50.0
& 49.8
& 49.8
\\

\bottomrule
\end{tabular}%
}
\caption{
Test PR-AUC for post-recognition attribution decisions, multiplied by 100.
Bold and underlined values indicate the best and second-best scores, respectively, in each column.
}
\label{tab:factual_side_probe_results}
\end{table*}

\Cref{tab:image_side_probe_results,tab:factual_side_probe_results} show that the most informative representation varies across attribution tasks. Image-side decisions are best captured by visual-token representations, whereas post-recognition decisions rely more on prompt hidden states. For entity recognition, \textsc{Vis} obtains the best average PR-AUC of 86.7, and for visual-evidence attribution, it reaches 97.0, far above the next best pre-generation feature, \textsc{Attn+Last8} at 67.6. The large gap indicates that the controlled recognizability boundary is strongly reflected in visual-token representations. This result should not be interpreted as evidence that naturally occurring visual failures are solved. Within this controlled stress test, visual-token representations are the most informative features.

The pattern changes after the entity has been recognized. For answer-success prediction, \textsc{Last8} with a linear head performs best for every model, reaching 62.7 average PR-AUC. This exceeds both \textsc{Vis} at 36.0 and the strongest post-generation baseline, \textsc{OutSeq} with a Transformer head, at 56.2. This suggests that prompt hidden states contain useful information about whether the model will answer correctly before decoding begins. Factual-access attribution is substantially harder. \textsc{Last8} with a Transformer head achieves the best average pre-generation result of 53.5, only slightly above \textsc{Oracle-UQ} at 52.4 (detailed results for UQ methods is in Appendix \ref{app:uq_results}). Answer-level uncertainty is also strongest for several individual models. We therefore interpret this result as evidence of a weak but useful pre-generation signal, rather than a reliable separation between \textsc{unknown fact} and \textsc{unrecallable fact}.

Image-side decisions also transfer more strongly from PopVQA to iNaturalist than post-recognition decisions; full cross-dataset results are provided in Appendix~\ref{app:cross-dataset}.

\paragraph{Does tree-structured composition help?}
A natural alternative to the tree-structured classifier is a 5-label classifier over the same outcome space.
For the tree-composed setting, we select the best-performing local probe for each decision from \Cref{tab:image_side_probe_results,tab:factual_side_probe_results} and compose their probabilities into five leaf-outcome scores using the attribution tree. For the flat setting, a single classifier directly predicts one of the five outcomes using its best validation-selected configuration. Hyperparameters are described in Appendix~\ref{app:grid-search}.

As shown in \Cref{tab:tree_vs_flat}, the direct 5-label classifier learns useful attribution signals but is consistently weaker than tree-structured composition, reaching 41.2 macro PR-AUC on average compared with 49.2. The difference appears across all four model, ranging from 5.1 points for Qwen2.5 to 11.4 points for Qwen3. This comparison does not establish the attribution tree as a unique causal decomposition. Rather, it shows that modeling the operational decisions separately is more effective than predicting all five outcomes from a single representation. The tree allows image-side decisions to rely on visual-token features and post-recognition decisions to rely on prompt-boundary hidden states, instead of forcing all outcomes through one representational bottleneck. Confusion matrices for the direct 5-label classifier are provided in Appendix~\ref{app:confusion-mat}.

\begin{table}[t]
\centering
\small
\setlength{\tabcolsep}{4pt}
\renewcommand{\arraystretch}{1.05}
\resizebox{\columnwidth}{!}{%
\begin{tabular}{lccccc}
\toprule
\textbf{Setting}
& \textbf{Gemma} & \textbf{Qwen2.5} & \textbf{Qwen3} & \textbf{Llama} & \textbf{Avg.} \\
\midrule
Tree-composed 
& \textbf{49.5} & \textbf{47.8} & \textbf{51.2} & \textbf{48.3} & \textbf{49.2} \\

Flat 5-label
& 40.9 & 42.7 & 39.8 & 41.5 & 41.2 \\
\bottomrule
\end{tabular}%
}
\caption{Comparison between attribution-tree composition and direct 5-label classification. Scores are test macro PR-AUC multiplied by 100.}
\label{tab:tree_vs_flat}
\end{table}

\section{Attribution-Guided Intervention}

The previous experiments show that pre-generation probes contain useful
signals for predicting operational attribution outcomes, extending recent evidence that pre-generation VLM states can support early abstention or routing~\citep{kogilathota-etal-2026-halp}. We test whether these predictions can guide targeted interventions. Given an image--question pair, the attribution tree predicts the most likely leaf outcome. If it predicts \textsc{success}, we keep the original answer; otherwise, the predicted failure type determines the intervention. For each predicted failure outcome, GPT-5 executes the corresponding intervention selected by the attribution tree; it does not determine which branch to apply. The interventions are as follows:
\begin{compactitem}
    \item Under \textsc{visual-evidence failure}, GPT-5 repairs the degraded image through image generation. The repaired image replaces the original, and the target VLM is rerun with the original question.

    \item Under \textsc{unknown entity}, GPT-5 identifies and briefly describes the entity. This information is appended to the original question before rerunning the target VLM.

    \item Under \textsc{unrecallable fact}, no external factual evidence is added. We first ask the target VLM to identify the entity,\footnote{For PopVQA, we ask ``What is the entity in the photo? Mention the proper name of the person/place/brand.'' For iNaturalist, we ask ``What is the entity in the photo? Mention the common name of the species.''} GPT-5 then rewrites the question to state the predicted entity explicitly, without adding factual information or answering the question. The target VLM is rerun on the rewritten question.

    \item Under \textsc{unknown fact}, GPT-5 provides short factual evidence relevant to the question and predicted entity. This evidence is appended to the original question before rerunning the target VLM.
\end{compactitem}

The exact prompts are provided in Appendix~\ref{app:mitigation-prompts}. In all cases, the attribution tree selects the intervention, while the target VLM generates the final answer. GPT-5 serves only as a general-purpose tool executor; deployment could instead use cheaper, task-specific tools.

\begin{table}[t]
\centering
\small
\setlength{\tabcolsep}{5pt}
\renewcommand{\arraystretch}{1.02}
\begin{tabular}{llccc}
\toprule
\textbf{Dataset} & \textbf{Model}
& \textbf{Before}
& \textbf{After}
& \textbf{Gain} \\
\midrule
\multirow{4}{*}{iNat}
& Gemma   & 10.1 & 46.6 & \textbf{+36.5} \\
& Qwen2.5 & 4.8  & 39.6 & \textbf{+34.8} \\
& Qwen3   & 6.3  & 45.6 & \textbf{+39.3} \\
& Llama   & 12.1 & 45.5 & \textbf{+33.4} \\
\midrule
\multirow{4}{*}{PopVQA}
& Gemma   & 31.0 & 63.9 & \textbf{+32.9} \\
& Qwen2.5 & 31.5 & 62.3 & \textbf{+30.8} \\
& Qwen3   & 30.1 & 64.3 & \textbf{+34.2} \\
& Llama   & 21.9 & 55.1 & \textbf{+33.2} \\
\bottomrule
\end{tabular}
\caption{Final answer accuracy (\%) before and after attribution-guided intervention. Gain is the absolute improvement in accuracy after mitigation.}
\label{tab:probe_guided_mitigation}
\end{table}

\begin{figure}[t]
\centering
\includegraphics[width=\columnwidth]{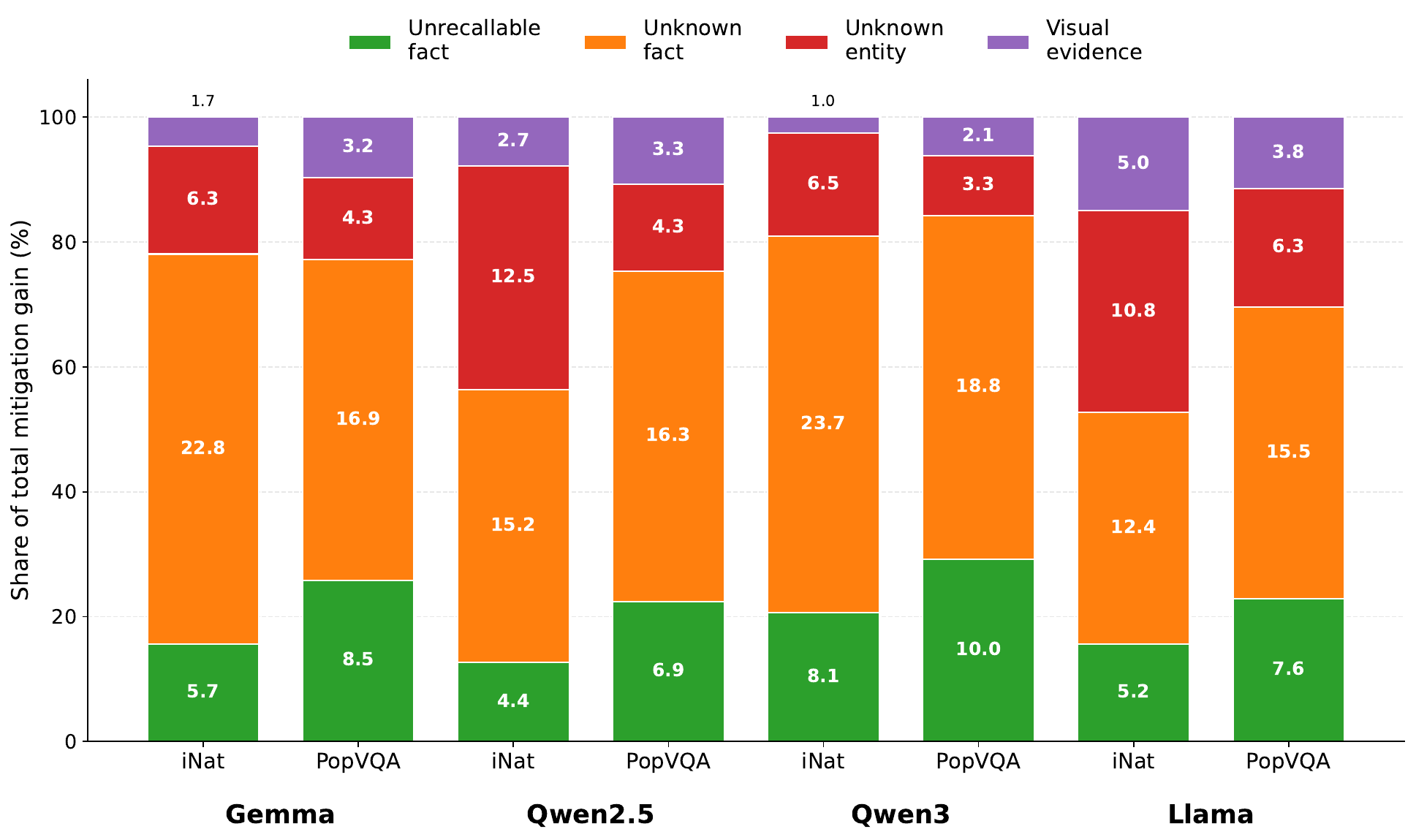}
\caption{Composition of attribution-guided intervention gains. Each bar shows how the total accuracy gain is distributed across failure modes for a given model and dataset. Bars are normalized to 100\%, while the numbers inside segments show the absolute percentage-point accuracy gain contributed by each failure mode.}
\label{fig:mitigation_gain_composition}
\end{figure}

Table~\ref{tab:probe_guided_mitigation} shows gains of
30.8--39.3 percentage points across all model--dataset pairs.
Figure~\ref{fig:mitigation_gain_composition} shows that
\textsc{unknown fact} contributes the largest share in most
settings, although the remaining branches also contribute.
Notably, \textsc{unrecallable fact} adds no external factual
evidence, while \textsc{visual-evidence failure} provides only a
repaired image. These results illustrate the practical benefits of routing
different predicted outcomes to distinct forms of intervention.


\section{Conclusion}

We studied knowledge-intensive VQA errors through an operational attribution tree over visual evidence, entity recognition, answer success, and factual access. Image-side decisions are best captured by visual-token representations, whereas post-recognition decisions are better supported by prompt-boundary hidden states. Factual-access attribution remains difficult, but exhibits useful signal for routing. These findings show that pre-generation representations can support fine-grained diagnostics and targeted interventions beyond binary failure detection or abstention.

\section*{Limitations}

Our framework relies on operational labels rather than direct human annotation of the true failure process. Recognition is measured with yes/no probes and distractors, factual access with entity-explicit rewrites, and visual-evidence failure with synthetic degradation. These procedures make the taxonomy scalable, but they may introduce noise and may not cover naturally occurring visual failures such as occlusion, unusual viewpoints, cropping, low resolution, or domain-specific artifacts.

The attribution labels are also tied to the target model and labeling protocol. For example, an entity may pass the recognition check while still being represented too coarsely for a downstream factual question, and entity-explicit rewrites may not fully separate missing knowledge from failed recall. The labels should therefore be interpreted as operational diagnostics rather than ground-truth causal annotations.

Finally, the mitigation experiment is a proof of concept rather than an optimized repair system. We use GPT-5 as a general-purpose support tool for entity support, factual evidence, question rewriting, and image repair, but this introduces additional cost and possible tool errors. Future work should evaluate cheaper specialized tools, naturally occurring visual failures, and per-example recovery under realistic deployment constraints.


\bibliography{custom}
\appendix

\section{Per-Layer Results}
\label{app:per-layer}

We analyze whether failure-attribution signals are localized in specific decoder layers or distributed across multiple late layers. Instead of concatenating the late-layer offsets used in the main experiments, we extract features from one decoder layer at a time. We denote these variants as \textsc{Vis}-L\(k\), \textsc{EOP}-L\(k\), \textsc{Last8}-L\(k\), \textsc{Attn}-L\(k\), and \textsc{Attn+Last8}-L\(k\), where \(k \in \{1,4,8,12\}\) is the offset below the final decoder layer. For example, \textsc{Last8}-L8 uses the final-eight prompt-token hidden states from the eighth layer below the final decoder layer.

\Cref{tab:per_layer_attribution_tree} reports the per-layer results for the four local attribution-tree decisions. Across most decisions and models, L8 is the strongest or near-strongest single layer, suggesting that useful failure-attribution signals are concentrated in mid-to-late decoder layers. However, the single-layer results remain below the corresponding multi-layer concatenated features in \Cref{tab:image_side_probe_results,tab:factual_side_probe_results}, indicating that the signal is distributed across late layers rather than fully localized to one layer. The representation preferences are consistent with the main results: visual-token features are strongest for image-side decisions, while prompt-boundary hidden states are strongest for downstream factual decisions.

\begin{table*}[ht]
\centering
\renewcommand{\arraystretch}{0.70}
\resizebox{\textwidth}{!}{%
\begin{tabular}{llccccc|ccccc}
\toprule
& & \multicolumn{5}{c|}{\textbf{Recognition:} recognized vs.\ not recognized} 
& \multicolumn{5}{c}{\textbf{Visual evidence:} unknown entity vs.\ visual-evidence failure} \\
\cmidrule(lr){3-7} \cmidrule(l){8-12}
\textbf{Feature} & \textbf{Head}
& \textbf{Gemma} & \textbf{Qwen2.5} & \textbf{Qwen3} & \textbf{Llama} & \textbf{Avg.}
& \textbf{Gemma} & \textbf{Qwen2.5} & \textbf{Qwen3} & \textbf{Llama} & \textbf{Avg.} \\
\midrule
\multicolumn{12}{l}{\textit{Single-layer visual-token features}} \\
\midrule
\textsc{Vis}-L1  & Lin. & 84.2 & 87.9 & 78.0 & 82.8 & 83.2 & 95.2 & 96.7 & 86.5 & 95.8 & 93.5 \\
\textsc{Vis}-L4  & Lin. & 85.7 & 89.5 & 79.6 & 84.2 & 84.8 & 96.6 & 98.1 & 88.0 & 97.4 & 95.0 \\
\textsc{Vis}-L8  & Lin. & 86.4 & 90.1 & 80.2 & 84.8 & 85.4 & 97.5 & 98.8 & 89.1 & 98.2 & 95.9 \\
\textsc{Vis}-L12 & Lin. & 85.1 & 88.7 & 78.9 & 83.6 & 84.1 & 96.1 & 97.5 & 87.4 & 96.8 & 94.5 \\

\midrule
\multicolumn{12}{l}{\textit{Single-layer hidden-state features}} \\
\midrule
\textsc{EOP}-L1  & Lin. & 78.9 & 67.8 & 81.5 & 68.9 & 74.3 & 57.8 & 56.3 & 57.2 & 58.6 & 57.5 \\
\textsc{EOP}-L4  & Lin. & 80.4 & 69.5 & 83.0 & 70.5 & 75.8 & 59.4 & 58.0 & 59.1 & 60.2 & 59.2 \\
\textsc{EOP}-L8  & Lin. & 81.6 & 71.0 & 84.2 & 71.6 & 77.1 & 60.8 & 59.2 & 60.5 & 61.4 & 60.5 \\
\textsc{EOP}-L12 & Lin. & 80.7 & 70.2 & 83.4 & 70.9 & 76.3 & 60.0 & 58.6 & 59.7 & 60.7 & 59.8 \\

\textsc{Last8}-L1  & Lin. & 79.8 & 71.2 & 80.9 & 68.7 & 75.2 & 62.8 & 60.6 & 55.7 & 66.5 & 61.4 \\
\textsc{Last8}-L4  & Lin. & 81.6 & 73.8 & 82.7 & 70.5 & 77.2 & 64.9 & 63.0 & 57.4 & 68.8 & 63.5 \\
\textsc{Last8}-L8  & Lin. & 82.7 & 75.0 & 84.0 & 71.8 & 78.4 & 66.1 & 64.4 & 58.8 & 70.2 & 64.9 \\
\textsc{Last8}-L12 & Lin. & 81.4 & 73.1 & 82.9 & 70.9 & 77.1 & 64.5 & 62.5 & 57.9 & 68.9 & 63.5 \\

\textsc{Last8}-L1  & Tr. & 80.2 & 71.5 & 81.8 & 66.9 & 75.1 & 62.0 & 60.0 & 56.8 & 65.4 & 61.0 \\
\textsc{Last8}-L4  & Tr. & 82.0 & 73.6 & 83.7 & 68.5 & 77.0 & 64.2 & 62.1 & 58.4 & 67.6 & 63.1 \\
\textsc{Last8}-L8  & Tr. & 83.1 & 75.0 & 85.0 & 70.0 & 78.3 & 65.4 & 63.6 & 60.0 & 69.2 & 64.5 \\
\textsc{Last8}-L12 & Tr. & 81.8 & 74.2 & 84.1 & 69.2 & 77.3 & 64.8 & 62.9 & 59.2 & 68.3 & 63.8 \\

\midrule
\multicolumn{12}{l}{\textit{Single-layer attention features}} \\
\midrule
\textsc{Attn}-L1  & Lin. & 81.2 & 70.4 & 81.8 & 68.4 & 75.5 & 61.5 & 60.8 & 61.7 & 63.4 & 61.9 \\
\textsc{Attn}-L4  & Lin. & 83.4 & 72.1 & 83.2 & 70.1 & 77.2 & 64.2 & 62.7 & 63.9 & 65.8 & 64.2 \\
\textsc{Attn}-L8  & Lin. & 84.6 & 73.4 & 84.3 & 71.4 & 78.4 & 65.5 & 64.1 & 64.9 & 67.2 & 65.4 \\
\textsc{Attn}-L12 & Lin. & 83.1 & 72.0 & 83.0 & 70.2 & 77.1 & 64.0 & 62.3 & 64.2 & 66.0 & 64.1 \\

\midrule
\multicolumn{12}{l}{\textit{Single-layer attention + hidden features}} \\
\midrule
\textsc{Attn+Last8}-L1  & Lin. & 79.5 & 73.2 & 82.4 & 68.8 & 76.0 & 63.4 & 61.9 & 58.4 & 67.6 & 62.8 \\
\textsc{Attn+Last8}-L4  & Lin. & 81.6 & 75.2 & 84.7 & 70.0 & 77.9 & 65.6 & 64.0 & 60.2 & 69.8 & 64.9 \\
\textsc{Attn+Last8}-L8  & Lin. & 83.0 & 76.5 & 86.0 & 71.4 & 79.2 & 67.0 & 65.8 & 61.4 & 71.4 & 66.4 \\
\textsc{Attn+Last8}-L12 & Lin. & 81.9 & 75.7 & 85.3 & 70.8 & 78.4 & 66.2 & 65.0 & 60.8 & 70.6 & 65.7 \\

\toprule
& & \multicolumn{5}{c|}{\textbf{Answer success:} success vs.\ failure after recognition} 
& \multicolumn{5}{c}{\textbf{Factual access:} unknown fact vs.\ unrecallable fact} \\
\cmidrule(lr){3-7} \cmidrule(l){8-12}
\textbf{Feature} & \textbf{Head}
& \textbf{Gemma} & \textbf{Qwen2.5} & \textbf{Qwen3} & \textbf{Llama} & \textbf{Avg.}
& \textbf{Gemma} & \textbf{Qwen2.5} & \textbf{Qwen3} & \textbf{Llama} & \textbf{Avg.} \\
\midrule
\multicolumn{12}{l}{\textit{Single-layer visual-token features}} \\
\midrule
\textsc{Vis}-L1  & Lin. & 32.4 & 33.6 & 31.2 & 32.0 & 32.3 & 30.5 & 29.4 & 28.0 & 31.8 & 29.9 \\
\textsc{Vis}-L4  & Lin. & 33.8 & 35.0 & 32.4 & 33.4 & 33.6 & 32.0 & 30.8 & 29.5 & 33.4 & 31.4 \\
\textsc{Vis}-L8  & Lin. & 34.8 & 36.2 & 33.6 & 34.6 & 34.8 & 33.1 & 31.8 & 30.4 & 34.5 & 32.5 \\
\textsc{Vis}-L12 & Lin. & 33.6 & 34.8 & 32.0 & 33.1 & 33.4 & 31.7 & 30.5 & 29.1 & 33.0 & 31.1 \\

\midrule
\multicolumn{12}{l}{\textit{Single-layer hidden-state features}} \\
\midrule
\textsc{EOP}-L1  & Lin. & 48.8 & 49.5 & 53.1 & 47.7 & 49.8 & 43.0 & 41.2 & 46.0 & 42.7 & 43.2 \\
\textsc{EOP}-L4  & Lin. & 50.6 & 51.5 & 55.0 & 49.5 & 51.6 & 44.8 & 42.7 & 47.9 & 44.3 & 44.9 \\
\textsc{EOP}-L8  & Lin. & 51.8 & 52.8 & 56.4 & 50.5 & 52.9 & 46.1 & 44.0 & 49.2 & 45.6 & 46.2 \\
\textsc{EOP}-L12 & Lin. & 50.9 & 52.0 & 55.3 & 49.9 & 52.0 & 45.3 & 43.4 & 48.4 & 44.9 & 45.5 \\

\textsc{Last8}-L1  & Lin. & 55.4 & 56.2 & 60.1 & 54.7 & 56.6 & 47.8 & 45.7 & 51.0 & 47.6 & 48.0 \\
\textsc{Last8}-L4  & Lin. & 58.3 & 59.1 & 63.2 & 57.0 & 59.4 & 50.1 & 47.8 & 53.4 & 49.6 & 50.2 \\
\textsc{Last8}-L8  & Lin. & 60.0 & 61.1 & 65.0 & 58.8 & 61.2 & 51.5 & 49.2 & 54.8 & 50.8 & 51.6 \\
\textsc{Last8}-L12 & Lin. & 58.7 & 59.7 & 63.8 & 57.5 & 59.9 & 50.6 & 48.1 & 53.9 & 50.0 & 50.6 \\

\textsc{Last8}-L1  & Tr. & 50.8 & 51.9 & 49.4 & 48.2 & 50.1 & 48.7 & 46.2 & 52.5 & 48.1 & 48.9 \\
\textsc{Last8}-L4  & Tr. & 52.9 & 54.1 & 51.5 & 50.2 & 52.2 & 50.7 & 48.1 & 54.5 & 49.8 & 50.8 \\
\textsc{Last8}-L8  & Tr. & 54.6 & 55.5 & 53.0 & 51.6 & 53.7 & 52.4 & 49.8 & 56.0 & 51.8 & 52.5 \\
\textsc{Last8}-L12 & Tr. & 53.5 & 54.7 & 52.1 & 50.8 & 52.8 & 51.6 & 49.1 & 55.2 & 51.0 & 51.7 \\

\midrule
\multicolumn{12}{l}{\textit{Single-layer attention features}} \\
\midrule
\textsc{Attn}-L1  & Lin. & 44.9 & 45.6 & 48.6 & 43.5 & 45.6 & 39.8 & 40.2 & 43.7 & 38.6 & 40.6 \\
\textsc{Attn}-L4  & Lin. & 47.2 & 48.0 & 50.8 & 45.8 & 48.0 & 42.1 & 42.7 & 46.1 & 40.5 & 42.9 \\
\textsc{Attn}-L8  & Lin. & 48.6 & 49.2 & 52.1 & 47.0 & 49.2 & 43.2 & 43.8 & 47.2 & 41.8 & 44.0 \\
\textsc{Attn}-L12 & Lin. & 47.4 & 48.2 & 51.0 & 46.0 & 48.1 & 42.5 & 43.0 & 46.3 & 40.9 & 43.2 \\

\midrule
\multicolumn{12}{l}{\textit{Single-layer attention + hidden features}} \\
\midrule
\textsc{Attn+Last8}-L1  & Lin. & 50.7 & 51.4 & 54.8 & 49.2 & 51.5 & 44.9 & 44.1 & 48.4 & 43.3 & 45.2 \\
\textsc{Attn+Last8}-L4  & Lin. & 52.6 & 53.4 & 56.7 & 51.0 & 53.4 & 46.7 & 45.8 & 50.2 & 45.0 & 46.9 \\
\textsc{Attn+Last8}-L8  & Lin. & 53.7 & 54.5 & 57.8 & 52.0 & 54.5 & 47.8 & 46.9 & 51.5 & 46.2 & 48.1 \\
\textsc{Attn+Last8}-L12 & Lin. & 52.9 & 53.8 & 57.0 & 51.3 & 53.8 & 47.0 & 46.2 & 50.7 & 45.5 & 47.4 \\
\bottomrule
\end{tabular}%
}
\caption{Per-layer performance for pre-generation features across the four attribution-tree binary decisions. Values are PR-AUC multiplied by 100. \(-Lk\) denotes extraction from the decoder layer \(k\) positions below the final layer.}
\label{tab:per_layer_attribution_tree}
\end{table*}
\section{LLM Judge Prompt for Factual Answers}
\label{app:eval_judge}

We use a text-only LLM (Meta-Llama-3-8B-Instruct\footnote{\url{https://huggingface.co/meta-llama/Meta-Llama-3-8B-Instruct}}) as a judge for short-form factual evaluation when exact string matching is insufficient (e.g., full-sentence answers, minor lexical variation). The judge is instructed to focus on the core factual content and accept minor lexical variants when they match any reference option.

\begin{tcolorbox}[
  colback=white,
  colframe=black,
  boxrule=0.6pt,
  arc=1.2mm,
  left=1.5mm,right=1.5mm,top=1.2mm,bottom=1.2mm,
  title=\textbf{LLM-judge prompt}
]
\small
\begin{Verbatim}[breaklines=true,breakanywhere=true]
You are a grader for short visual QA. The ground truth may include multiple acceptable answers separated by '|'.
The model answer may be a full sentence; judge based on the core factual answer.
Treat minor spelling or adjective variations as correct if the meaning matches any reference option.
Respond with only one word: Correct or Incorrect.

Question: {question}
Ground truth options: {reference}
Model answer: {candidate}

Does the model answer match any ground truth option? Respond with Correct or Incorrect.
\end{Verbatim}
\end{tcolorbox}

\section{Representative Entity-Linked Factual Questions}
\label{app:sample-qa}

The following examples illustrate the form of entity-linked factual questions in our evaluation sets. Each item is associated with an image depicting the named entity (species, person, place, or artwork). We show the question and the reference answer used for evaluation.

\begin{tcolorbox}[colback=white,colframe=black,boxrule=0.5pt,arc=1mm,
  left=1mm,right=1mm,top=0.8mm,bottom=0.8mm,title=\textbf{iNaturalist}]
\small
\begin{itemize}
  \item \textbf{West Indian Fuzzy Chiton.} \emph{What is the habitat of this animal?} \ $\rightarrow$ \textit{rocks very high in the intertidal zone}
  \item \textbf{Horse Lubber Grasshopper.} \emph{Besides chemical deterrents, what else does this insect use to ward off predators?} \ $\rightarrow$ \textit{visual and auditory elements}
  \item \textbf{Sweet cherry.} \emph{What is the height range of this tree?} \ $\rightarrow$ \textit{15-32 metres}
  \item \textbf{Temple Tree Frog.} \emph{In which part of the world does this animal live?} \ $\rightarrow$ \textit{Taiwan}
\end{itemize}
\end{tcolorbox}

\begin{tcolorbox}[colback=white,colframe=black,boxrule=0.5pt,arc=1mm,
  left=1mm,right=1mm,top=0.8mm,bottom=0.8mm,title=\textbf{PopVQA}]
\small
\begin{itemize}
  \item \textbf{Gerasimov Institute of Cinematography.} \emph{In what country is the place in this image located?} \ $\rightarrow$ \textit{Russia}
  \item \textbf{Hunter Hayes.} \emph{What instrument does the subject of this image play?} \ $\rightarrow$ \textit{guitar}
  \item \textbf{Arkadi Monastery.} \emph{What is the architectural style of the place in this image?} \ $\rightarrow$ \textit{baroque architecture}
  \item \textbf{Girl in a Blue Dress.} \emph{A part of what collection is the painting in this image?} \ $\rightarrow$ \textit{The Wallace Collection}
\end{itemize}
\end{tcolorbox}

\section{Formalization of Pre-Generation Feature Extraction and Training}
\label{app:rq2-formal}

\paragraph{Notation.}
Let $x$ denote the input image and $q$ the textual question. The model is a vision-language transformer that maps the multimodal prompt into a sequence of hidden states over $T$ prompt tokens. We write the prompt token sequence as
$\mathbf{z}_{1:T} = \mathrm{Tok}(x,q)$,
where the sequence contains both visual prompt tokens and text tokens. Let $L$ be the number of transformer layers and let $\ell \in \{1,\dots,L\}$ index layers, with $\ell=L$ being the last layer. We use negative offsets for convenience: layer $-k$ refers to $\ell=L-k+1$.

We denote by $\mathbf{h}_t^{(\ell)} \in \mathbb{R}^d$ the hidden state at prompt token position $t$ and layer $\ell$, after processing the full prompt but \emph{before} decoding any answer token. We denote attention weights by
$\alpha_{t,s}^{(\ell, m)} \in [0,1]$,
the attention from token $t$ to token $s$ at layer $\ell$ and head $m\in\{1,\dots,M\}$, with $\sum_{s=1}^{T}\alpha_{t,s}^{(\ell,m)}=1$.

\paragraph{Prediction targets.}
For each local decision $d\in\{R,D,S,F\}$, let
$y_i^{(d)}\in\{0,1\}$ denote the corresponding operational label:
entity recognition for $R$, visual-evidence attribution for $D$,
answer success for $S$, and factual access for $F$. Each probe is
trained only on examples reaching its decision node in the attribution
tree.

\paragraph{Feature families.}
We extract features from a single forward pass over the prompt.

\subparagraph{(1) Visual-prompt features.}
Let \(t_{\mathrm{img}}\) be the final image-token position. At layer
\(\ell\), the visual-prompt feature is
\begin{equation}
\boldsymbol{\phi}_{\mathrm{vis}}^{(\ell)}(x,q)
=
\mathbf{h}_{t_{\mathrm{img}}}^{(\ell)}.
\label{eq:vis_last_image}
\end{equation}
For a set of layers \(\mathcal{L}\), we concatenate these
representations:
\[
\boldsymbol{\phi}_{\mathrm{vis}}^{(\mathcal{L})}
=
\mathrm{Concat}_{\ell\in\mathcal{L}}
\boldsymbol{\phi}_{\mathrm{vis}}^{(\ell)}.
\]

\subparagraph{(2) Hidden-state features.}
Let \(t_{\mathrm{eop}}\) denote the final prompt-token position. The
end-of-prompt feature at layer \(\ell\) is
\begin{equation}
\boldsymbol{\phi}_{\mathrm{eop}}^{(\ell)}(x,q)
=
\mathbf{h}_{t_{\mathrm{eop}}}^{(\ell)}.
\label{eq:eop_hidden}
\end{equation}
For \textsc{Last8}, let
\(\mathcal{T}_8=\{t_{\mathrm{eop}}-7,\ldots,t_{\mathrm{eop}}\}\)
denote the final eight prompt-token positions, truncated at the
beginning of the sequence when necessary. We define
\begin{equation}
\boldsymbol{\phi}_{\mathrm{last8}}^{(\ell)}(x,q)
=
\mathrm{Vec}\!\left(
\left[\mathbf{h}_{t}^{(\ell)}\right]_{t\in\mathcal{T}_8}
\right)
\in\mathbb{R}^{8d},
\label{eq:last8_hidden}
\end{equation}
where \(\mathrm{Vec}(\cdot)\) flattens the \(8\times d\) matrix.
Multi-layer variants concatenate the corresponding representations:
\[
\boldsymbol{\phi}_{a}^{(\mathcal{L})}
=
\mathrm{Concat}_{\ell\in\mathcal{L}}
\boldsymbol{\phi}_{a}^{(\ell)},
\qquad
a\in\{\mathrm{eop},\mathrm{last8}\}.
\]

\subparagraph{(3) Attention-map features with a finite lookback.}
For each $t\in\mathcal{T}_8$, we restrict attention to a short history window of size $k$ (here $k=4$):
\begin{equation}
\mathcal{H}_k(t)=\{\,s \mid \max(1,t-k)\le s \le t-1\,\}.
\label{eq:hist_set}
\end{equation}
For a fixed layer $\ell$, we form attention features by collecting and flattening
$\alpha_{t,s}^{(\ell,m)}$ for $t\in\mathcal{T}_8$, $s\in\mathcal{H}_k(t)$, and $m\in\{1,\dots,M\}$:
\begin{align}
&\boldsymbol{\phi}_{\mathrm{attn}}^{(\ell)}(x,q)
=\nonumber \\
&\mathrm{Vec}\!\left(\left[\alpha_{t,s}^{(\ell,m)}\right]_{t\in\mathcal{T}_8,\ s\in\mathcal{H}_k(t),\ m=1}^{M}\right)\in\mathbb{R}^{8kM}.
\label{eq:attn_feat}
\end{align}
We then concatenate across layers $\mathcal{L}$. The combined hidden and attention representation is then
\begin{equation}
\boldsymbol{\phi}_{\mathrm{last8}}^{(\mathcal{L})}(x,q)\ \Vert\ 
\boldsymbol{\phi}_{\mathrm{attn}}^{(\mathcal{L})}(x,q),
\label{eq:hidden_attn_concat}
\end{equation}
where $\Vert$ denotes vector concatenation.

\subparagraph{(4) Post-generation baseline.}
For comparison, we also evaluate a decoding-dependent baseline that extracts features from the generated output sequence following~\cite{shelmanov-etal-2025-head}. Let the model generate an answer token sequence $\mathbf{y}_{1:N}$, and let $t_i$ denote the $i$-th generated token. Let $\alpha_{i,i-j}^{(\ell,m)}$ denote attention from token position $i$ to a previous position $(i-j)$ within the \emph{generated sequence} at layer $\ell$ and head $m$. For a small history window size $k$ and a chosen set of layers $\mathcal{L}$, the attention feature for token $t_i$ is
\begin{equation}
\mathbf{f}_{\mathrm{att}}(t_i)=
\mathrm{Vec}\!\left(\left[\alpha_{i,i-j}^{(\ell,m)}\right]_{j=1}^{k}\right)_{\ell\in\mathcal{L},\,m=1}^{M}
\label{eq:shel_att}
\end{equation}
Let $\mathrm{top}\text{-}m(P_i)$ denote the $m$ most probable tokens under the model's next-token distribution
$P_i(\cdot)=P(\cdot \mid x,q,\mathbf{y}_{<i})$.
The probability feature is
\begin{equation}
\mathbf{f}_{\mathrm{prob}}(t_i)=\Big[\log P_i(u)\Big]_{u\in \mathrm{top}\text{-}m(P_i)}
\label{eq:shel_prob}
\end{equation}
Then, the features are concatenated:
\begin{equation}
\mathbf{f}(t_i)=\mathbf{f}_{\mathrm{att}}(t_i)\ \Vert\ \mathbf{f}_{\mathrm{prob}}(t_i)
\label{eq:shel_concat}
\end{equation}
For the linear head, we mean-pool the token-level feature sequence to obtain a fixed-dimensional vector. For the Transformer head, we retain the sequence of token-level features and apply the sequence encoder described below. This baseline depends on decoding trajectories and next-token distributions, whereas our main approach uses prompt-boundary representations prior to generation.

\subparagraph{(5) Token probability feature.}
Finally, we include a lightweight post-generation baseline based only on the model's next-token probability distribution. Although token probabilities can be miscalibrated, they still provide useful information about the model's conditional confidence at each decoding step. Let \(P_i(\cdot)=P(\cdot \mid x,q,\mathbf{y}_{<i})\) be the next-token distribution before generating token \(y_i\). For each generated position \(i\), we collect the log-probabilities of the top-\(m\) candidate tokens:
\begin{equation}
\mathbf{f}_{\mathrm{prob}}(t_i)
=
\left\{
\log P_i(u)
\;\middle|\;
u \in \mathrm{top}\text{-}m(P_i)
\right\}.
\label{eq:tokprob_topm}
\end{equation}
We then concatenate these features across generated positions up to a fixed maximum length \(N_{\max}\):
\begin{equation}
\boldsymbol{\phi}_{\mathrm{prob}}(x,q,\mathbf{y})
=
\mathrm{Concat}_{i=1}^{N_{\max}}
\mathbf{f}_{\mathrm{prob}}(t_i),
\label{eq:tokprob_seq}
\end{equation}
with padding for shorter generations and truncation for longer ones. A linear classification head is trained on this concatenated vector.

\paragraph{Prediction heads and training.}
Given a feature representation $\boldsymbol{\phi}_i=\boldsymbol{\phi}(x_i,q_i)$, we use either a linear head or, for sequence-structured features, a Transformer-based head. For vector-valued features, the linear head produces logits
\begin{equation}
\mathbf{z}_i=\mathbf{W}\boldsymbol{\phi}_i+\mathbf{b},
\label{eq:linear_head}
\end{equation}
where $\mathbf{W}\in\mathbb{R}^{C\times p}$ and $\mathbf{b}\in\mathbb{R}^{C}$, with $C$ determined by the label space of the prediction task. For sequence-valued features $\mathbf{Z}_i\in\mathbb{R}^{T\times D_{\mathrm{in}}}$ with mask $\mathbf{m}_i\in\{0,1\}^T$, the Transformer head first applies layer normalization, a linear projection to $d_{\mathrm{model}}$, GELU, and dropout, prepends a learned \texttt{[CLS]} token, and then encodes the sequence with a Transformer encoder. The final \texttt{[CLS]} representation is passed to an MLP classifier to produce logits $\mathbf{z}_i\in\mathbb{R}^{C}$. In all current experiments, the Transformer head uses $d_{\mathrm{model}}=256$, $n_{\mathrm{heads}}=8$, $n_{\mathrm{layers}}=2$, and dropout $=0.1$. We optimize a cross-entropy objective appropriate to the task label space: binary cross-entropy for binary targets and multiclass cross-entropy for categorical targets. Hyperparameters are selected by grid search on the development split, and final performance is reported on a held-out test split.

\section{Supplementary Statistics}
\label{app:additional}

\subsection{Grid Search Hyperparameters}
\label{app:grid-search}

We tune each feature family, head type, and model independently on
the validation split. For the binary probes, configurations are
selected using validation PR-AUC for the corresponding attribution
decision. For the direct 5-label classifier, configurations are
selected using validation macro PR-AUC. The flat setting includes a
smaller learning rate, a smaller batch size, and a longer training
option to account for the multiclass optimization problem.
Table~\ref{tab:grid_search_hyperparameters} summarizes both search
spaces.

\begin{table}[H]
\centering
\scriptsize
\setlength{\tabcolsep}{3pt}
\renewcommand{\arraystretch}{1.08}
\begin{tabular}{lcc}
\toprule
\textbf{Hyperparameter}
& \textbf{Binary probes}
& \textbf{Flat classifier} \\
\midrule
Learning rate
& $\{10^{-4},\,3{\times}10^{-4},\,10^{-3}\}$
& \shortstack{$\{3{\times}10^{-5},\,10^{-4},$\\
              $3{\times}10^{-4},\,10^{-3}\}$} \\

Weight decay
& $\{0,\,10^{-5},\,10^{-4}\}$
& $\{0,\,10^{-5},\,10^{-4}\}$ \\

Batch size
& $\{128,\,256\}$
& $\{64,\,128,\,256\}$ \\

Epochs
& $\{5,\,10,\,15,\,20\}$
& $\{10,\,15,\,20,\,30\}$ \\
\bottomrule
\end{tabular}
\caption{Hyperparameter grids for the binary attribution probes and
the direct 5-label classifier.}
\label{tab:grid_search_hyperparameters}
\end{table}
\subsection{Validation Results}
\label{app:val-results-t2-t4}

\subsubsection{Validation Performance for Attribution-Tree Decisions}

\Cref{tab:image_side_val_results,tab:factual_side_val_results} report validation-set PR-AUC for the four local decisions in the attribution tree. These results are computed on the development split used for hyperparameter selection. The validation results follow the
same qualitative pattern as the held-out test results: \textsc{Vis} is strongest for image side decisions, particularly visual-evidence attribution, while \textsc{Last8} features are strongest for post-recognition decisions. Post-generation \textsc{OutSeq} features remain competitive for answer-success and factual-access prediction, but do not outperform the best pre-generation prompt-boundary representations on average.

\begin{table*}[t]
\centering
\small
\renewcommand{\arraystretch}{0.82}
\resizebox{\textwidth}{!}{%
\begin{tabular}{llccccc|ccccc}
\toprule
& &
\multicolumn{5}{c|}{
\textbf{Recognition:} recognized vs.\ unrecognized
}
&
\multicolumn{5}{c}{
\textbf{Visual evidence:} unknown entity vs.\ visual-evidence failure
}
\\
\cmidrule(lr){3-7}
\cmidrule(l){8-12}

\textbf{Method} & \textbf{Head}
& \textbf{Gemma}
& \textbf{Qwen2.5}
& \textbf{Qwen3}
& \textbf{Llama}
& \textbf{Avg.}
& \textbf{Gemma}
& \textbf{Qwen2.5}
& \textbf{Qwen3}
& \textbf{Llama}
& \textbf{Avg.}
\\
\midrule

\multicolumn{12}{l}{\textit{Pre-generation probes}} \\

\textsc{Vis} & Lin.
& \textbf{88.2}
& \textbf{91.7}
& 82.0
& \textbf{86.4}
& \textbf{87.1}
& \textbf{98.7}
& \textbf{99.8}
& \textbf{91.3}
& \textbf{99.4}
& \textbf{97.3}
\\

\textsc{EOP} & Lin.
& 82.6
& 72.5
& 85.6
& 72.9
& 78.4
& 62.4
& 60.7
& 62.1
& 63.1
& 62.1
\\

\textsc{Last8} & Lin.
& 84.2
& 76.2
& 85.5
& \underline{73.7}
& 79.9
& 67.7
& 66.1
& 60.4
& 71.5
& 66.4
\\

\textsc{Last8} & Tr.
& 84.5
& 76.8
& 86.5
& 71.5
& 79.8
& 67.1
& 65.3
& 61.8
& 70.6
& 66.2
\\

\textsc{Attn} & Lin.
& \underline{86.0}
& 74.7
& 85.6
& 73.1
& 79.9
& 67.2
& 65.9
& \underline{66.3}
& 68.8
& 67.1
\\

\textsc{Attn+Last8} & Lin.
& 84.2
& \underline{77.9}
& \textbf{87.4}
& 72.5
& \underline{80.5}
& \underline{68.9}
& \underline{67.4}
& 63.1
& \underline{72.9}
& \underline{68.1}
\\

\midrule
\multicolumn{12}{l}{\textit{Post-generation baselines}} \\

\textsc{OutSeq} & Lin.
& 80.9
& 76.0
& \underline{86.9}
& 65.5
& 77.3
& 56.6
& 59.9
& 54.8
& 54.2
& 56.4
\\

\textsc{OutSeq} & Tr.
& 83.5
& 73.9
& 85.2
& 68.3
& 77.7
& 58.0
& 61.2
& 56.2
& 55.6
& 57.8
\\

\bottomrule
\end{tabular}%
}
\caption{
Validation PR-AUC for image-side attribution decisions. Values are multiplied by 100. Bold indicates the best score in each model column; underlining indicates the second-best score.
}
\label{tab:image_side_val_results}
\end{table*}

\begin{table*}[ht]
\centering
\small
\renewcommand{\arraystretch}{0.82}
\resizebox{\textwidth}{!}{%
\begin{tabular}{llccccc|ccccc}
\toprule
& &
\multicolumn{5}{c|}{
\textbf{Answer success:} success vs.\ failure after recognition
}
&
\multicolumn{5}{c}{
\textbf{Factual access:} unknown fact vs.\ unrecallable fact
}
\\
\cmidrule(lr){3-7}
\cmidrule(l){8-12}

\textbf{Method} & \textbf{Head}
& \textbf{Gemma}
& \textbf{Qwen2.5}
& \textbf{Qwen3}
& \textbf{Llama}
& \textbf{Avg.}
& \textbf{Gemma}
& \textbf{Qwen2.5}
& \textbf{Qwen3}
& \textbf{Llama}
& \textbf{Avg.}
\\
\midrule

\multicolumn{12}{l}{\textit{Pre-generation probes}} \\

\textsc{Vis} & Lin.
& 36.6
& 37.9
& 35.2
& 36.1
& 36.5
& 34.3
& 32.9
& 31.7
& 35.6
& 33.6
\\

\textsc{EOP} & Lin.
& 53.3
& 54.6
& 58.3
& 52.1
& 54.6
& 47.7
& 45.7
& 50.7
& 47.4
& 47.9
\\

\textsc{Last8} & Lin.
& \textbf{61.8}
& \textbf{63.1}
& \textbf{67.3}
& \textbf{60.8}
& \textbf{63.2}
& \underline{52.6}
& \underline{50.4}
& \underline{56.2}
& \underline{51.8}
& \underline{52.7}
\\

\textsc{Last8} & Tr.
& 56.1
& 57.4
& 54.8
& \underline{53.6}
& 55.5
& \textbf{53.8}
& \textbf{51.3}
& \textbf{57.7}
& \textbf{53.1}
& \textbf{54.0}
\\

\textsc{Attn} & Lin.
& 50.2
& 50.9
& 54.0
& 48.7
& 50.9
& 44.3
& 45.0
& 48.7
& 43.4
& 45.3
\\

\textsc{Attn+Last8} & Lin.
& 54.9
& 55.7
& \underline{59.2}
& 53.4
& 55.8
& 49.0
& 48.5
& 52.8
& 47.7
& 49.5
\\

\midrule
\multicolumn{12}{l}{\textit{Post-generation baselines}} \\

\textsc{OutSeq} & Lin.
& 53.0
& 53.9
& 55.8
& 50.5
& 53.3
& 50.0
& 48.3
& 52.4
& 48.8
& 49.9
\\

\textsc{OutSeq} & Tr.
& \underline{56.4}
& \underline{58.0}
& 58.8
& 53.2
& \underline{56.6}
& 52.2
& 50.3
& 54.8
& 51.1
& 52.1
\\

\bottomrule
\end{tabular}%
}
\caption{
Validation PR-AUC for post-recognition attribution decisions. Values are multiplied by 100. Bold indicates the best score in each
model column; underlining indicates the second-best score.
}
\label{tab:factual_side_val_results}
\end{table*}

\subsubsection{Validation Performance for the Flat Setting}

\Cref{tab:flat_5label_val} reports validation macro PR-AUC for direct 5-label classification. Unlike the attribution-tree setting, the flat classifier must predict all final outcomes from a single feature representation. The strongest validation configuration is \textsc{Last8} with a Transformer head, reaching 43.0 average macro PR-AUC. \textsc{Last8} with a linear head is close behind at 42.0, while \textsc{Attn+Last8} and \textsc{EOP} are weaker.

\begin{table}[H]
\centering
\small
\setlength{\tabcolsep}{4pt}
\renewcommand{\arraystretch}{1.05}
\resizebox{\columnwidth}{!}{%
\begin{tabular}{llccccc}
\toprule
\textbf{Feature} & \textbf{Head}
& \textbf{Gemma} & \textbf{Qwen2.5} & \textbf{Qwen3} & \textbf{Llama} & \textbf{Avg.} \\
\midrule
\textsc{Vis} & Lin.
& 34.8 & 36.2 & 33.9 & 35.1 & 35.0 \\

\textsc{EOP} & Lin.
& 39.6 & 40.8 & 38.7 & 39.9 & 39.8 \\

\textsc{Last8} & Lin.
& 42.1 & 43.0 & 40.9 & 41.8 & 42.0 \\

\textsc{Last8} & Tr.
& \textbf{43.2} & \textbf{44.1} & \textbf{41.7} & \textbf{42.9} & \textbf{43.0} \\

\textsc{Attn} & Lin.
& 37.8 & 39.1 & 36.9 & 38.0 & 38.0 \\

\textsc{Attn+Last8} & Lin.
& 40.5 & 41.8 & 39.6 & 40.7 & 40.7 \\
\bottomrule
\end{tabular}%
}
\caption{Validation macro PR-AUC for direct 5-label classification. Scores are multiplied by 100.}
\label{tab:flat_5label_val}
\end{table}

\subsection{Confusion Matrices}
\label{app:confusion-mat}

\begin{figure}[t]
    \centering
    \includegraphics[width=\linewidth]{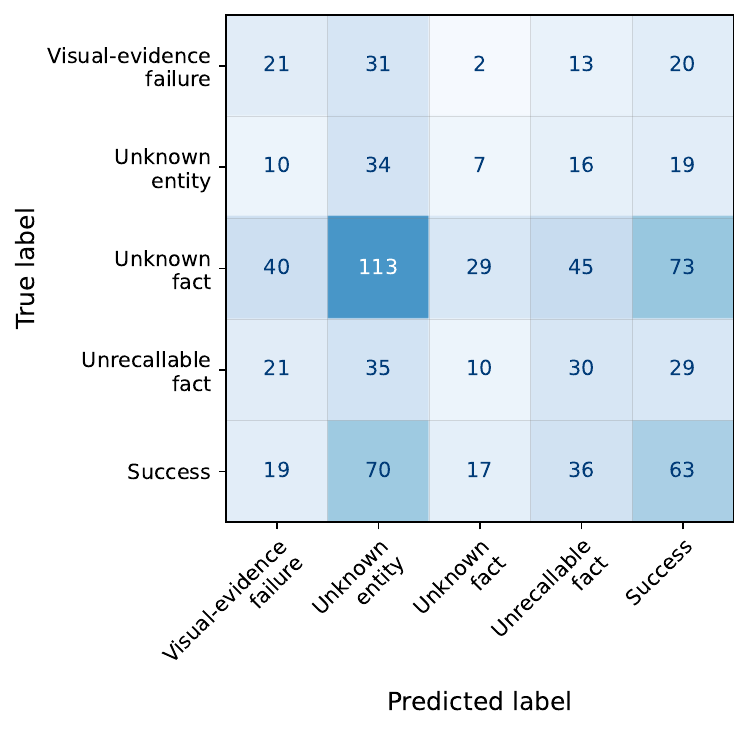}
    \caption{Confusion matrix for the direct 5-label classifier on Gemma. The classifier assigns many examples to \textsc{unknown entity} and \textsc{success}, while downstream factual classes remain difficult to separate.}
    \label{fig:confusion_gemma_5class}
\end{figure}

\begin{figure}[t]
    \centering
    \includegraphics[width=\linewidth]{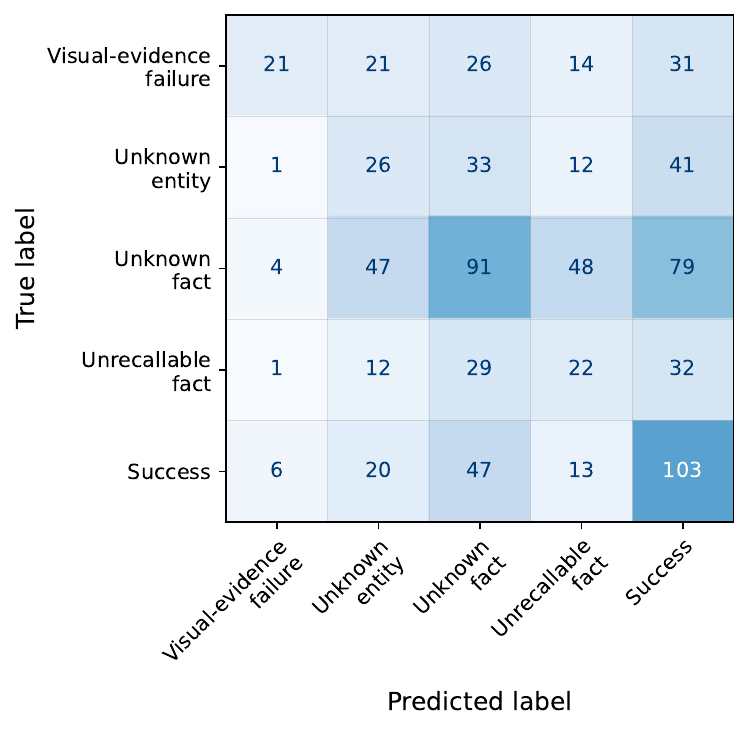}
    \caption{Confusion matrix for the direct 5-label classifier on Qwen2.5. Compared with the other models, predictions are more distributed across labels, but substantial confusion remains between \textsc{unknown fact}, \textsc{unrecallable fact}, and \textsc{success}.}
    \label{fig:confusion_qwen2p5_5class}
\end{figure}

\begin{figure}[t]
    \centering
    \includegraphics[width=\linewidth]{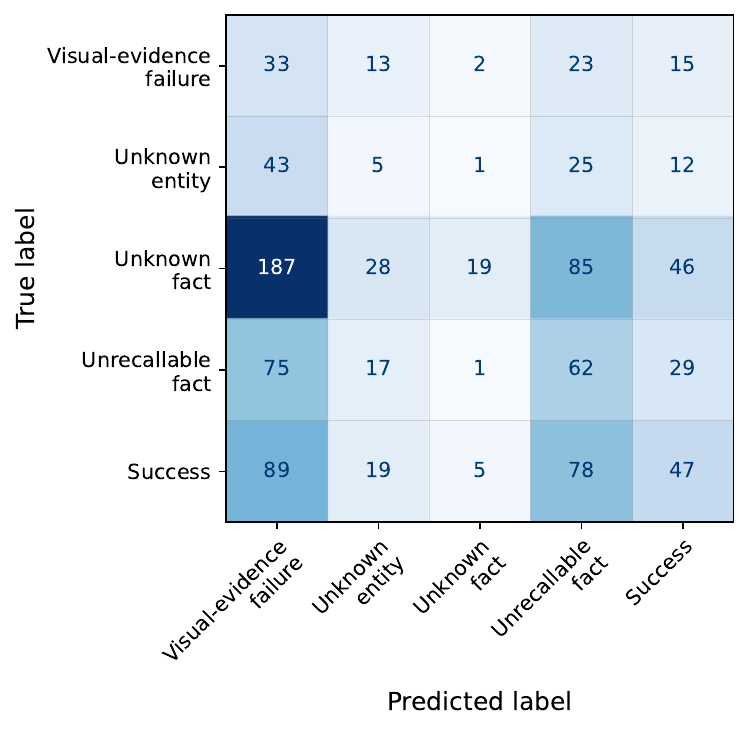}
    \caption{Confusion matrix for the direct 5-label classifier on Qwen3. The classifier frequently over-predicts \textsc{visual-evidence failure} and \textsc{unrecallable fact}.}
    \label{fig:confusion_qwen3_5class}
\end{figure}

\begin{figure}[t]
    \centering
    \includegraphics[width=\linewidth]{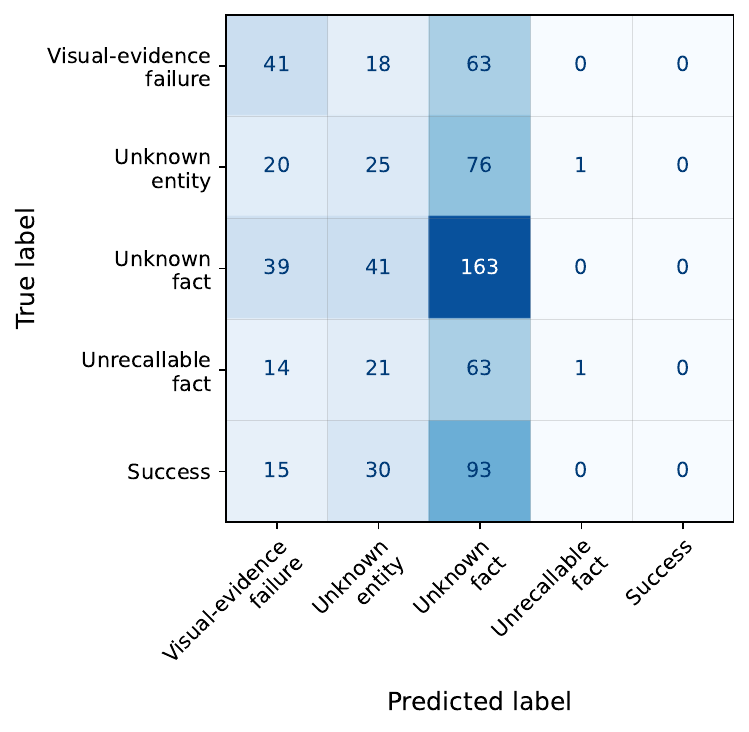}
    \caption{Confusion matrix for the direct 5-label classifier on Llama. Predictions collapse heavily toward \textsc{unknown fact}, with very few examples assigned to \textsc{unrecallable fact} or \textsc{success}.}
    \label{fig:confusion_llama_5class}
\end{figure}

To better understand the behavior of the direct 5-label classifier, we report confusion matrices in Figures~\ref{fig:confusion_gemma_5class},~\ref{fig:confusion_qwen2p5_5class},~\ref{fig:confusion_qwen3_5class},~\ref{fig:confusion_llama_5class} for each model. Rows correspond to the operational attribution label, and columns correspond to the predicted label. These plots show that the flat classifier often confuses neighboring or related attribution outcomes, especially downstream factual labels such as \textsc{unknown fact}, \textsc{unrecallable fact}, and \textsc{success}. This supports the motivation for the attribution-tree formulation: when all labels are predicted in a single step, the classifier must learn boundaries between image-side failures and later factual failures simultaneously, which can lead to systematic confusion.

The confusion matrices further illustrate why the direct 5-label classifier underperforms the attribution-tree formulation. Across models, the flat classifier does not simply make isolated mistakes between neighboring labels; instead, it often collapses toward a small subset of outcomes. Gemma assigns many examples to \textsc{unknown entity} and \textsc{success}, while Llama heavily over-predicts \textsc{unknown fact} and almost never predicts \textsc{success} or \textsc{unrecallable fact}. Qwen3 shows a different but related failure pattern, with many factual and success examples mapped to \textsc{visual-evidence failure} or \textsc{unrecallable fact}. Even Qwen2.5, whose predictions are more distributed, still confuses \textsc{unknown fact}, \textsc{unrecallable fact}, and \textsc{success}. These patterns show that the five outcomes are not equally separable from a single flat representation, which is consistent with these attribution decisions relying on
different internal signals.

\section{Performance of Uncertainty Quantification Methods}
\label{app:uq_results}

Table~\ref{tab:uq_full_results} reports the detailed PR-AUC results
for the uncertainty baselines across the four local decisions. Overall, these methods remain close to the corresponding random PR-AUC baselines. The strongest average improvements are 2.8 points for recognition and 1.9 points for visual-evidence attribution, while the gains for answer success and factual access are below one point.

\begin{table*}[ht]
\centering
\small
\setlength{\tabcolsep}{3.5pt}
\renewcommand{\arraystretch}{0.88}
\resizebox{\textwidth}{!}{%
\begin{tabular}{lccccc|ccccc}
\toprule
& \multicolumn{5}{c|}{\textbf{Recognition:} recognized vs.\ not recognized}
& \multicolumn{5}{c}{\textbf{Visual evidence:} unknown entity vs.\ visual-evidence failure} \\
\cmidrule(lr){2-6} \cmidrule(l){7-11}
\textbf{UQ method}
& \textbf{Gemma} & \textbf{Qwen2.5} & \textbf{Qwen3} & \textbf{Llama} & \textbf{Avg.}
& \textbf{Gemma} & \textbf{Qwen2.5} & \textbf{Qwen3} & \textbf{Llama} & \textbf{Avg.} \\
\midrule
\textsc{MSP}~\citep{hendrycks2017baseline}
& 80.6 & 72.6 & 85.1 & 70.4 & 77.2
& 37.3 & 36.4 & 30.0 & 29.0 & 33.2 \\

\textsc{MTE}~\citep{fomicheva-etal-2020-unsupervised}
& 80.3 & 73.6 & 84.9 & 70.2 & 77.3
& 35.3 & 36.8 & 29.7 & 29.3 & 32.7 \\

\textsc{LexSim}~\citep{fomicheva-etal-2020-unsupervised}
& 81.2 & 73.0 & 84.7 & 68.5 & 76.9
& 35.9 & 34.7 & 29.8 & 28.7 & 32.2 \\

\textsc{DegMat}~\citep{lin2024generating}
& 81.1 & 73.6 & 84.8 & 68.5 & 77.0
& 34.3 & 34.7 & 29.7 & 30.0 & 32.2 \\

\textsc{SemanticEntropy}~\citep{kuhn2023semantic}
& 80.5 & 73.7 & 84.6 & 70.2 & 77.3
& 36.2 & 35.5 & 30.0 & 29.7 & 32.9 \\

\textsc{SAR}~\citep{duan-etal-2024-shifting}
& 80.5 & 73.0 & 84.0 & 70.5 & 77.0
& 36.9 & 35.3 & 30.4 & 29.2 & 32.9 \\

\textsc{SemanticDensity}~\citep{qiu2024semanticdensity}
& 81.0 & 71.5 & 84.5 & 68.5 & 76.4
& 33.7 & 34.2 & 28.8 & 31.6 & 32.1 \\

\textsc{CoCoA-MSP}~\citep{vashurin2025cocoa}
& 80.3 & 73.3 & 84.8 & 69.9 & 77.1
& 37.3 & 36.1 & 30.2 & 29.6 & 33.3 \\

\textsc{CoCoA-PPL}~\citep{vashurin2025cocoa}
& 80.0 & 73.0 & 84.6 & 68.6 & 76.6
& 36.3 & 35.6 & 30.2 & 29.2 & 32.8 \\

\textsc{CoCoA-MTE}~\citep{vashurin2025cocoa}
& 79.9 & 73.8 & 84.6 & 68.8 & 76.8
& 36.1 & 36.1 & 30.1 & 29.3 & 32.9 \\

\midrule
Random PR-AUC
& 78.5 & 71.0 & 82.1 & 66.3 & 74.5
& 32.6 & 34.1 & 30.2 & 28.7 & 31.4 \\

\toprule
& \multicolumn{5}{c|}{\textbf{Answer success:} success vs.\ failure after recognition}
& \multicolumn{5}{c}{\textbf{Factual access:} unknown fact vs.\ unrecallable fact} \\
\cmidrule(lr){2-6} \cmidrule(l){7-11}
\textbf{UQ method}
& \textbf{Gemma} & \textbf{Qwen2.5} & \textbf{Qwen3} & \textbf{Llama} & \textbf{Avg.}
& \textbf{Gemma} & \textbf{Qwen2.5} & \textbf{Qwen3} & \textbf{Llama} & \textbf{Avg.} \\
\midrule
\textsc{MSP}~\citep{hendrycks2017baseline}
& 30.5 & 28.3 & 34.3 & 26.9 & 30.0
& 55.0 & 48.8 & 47.0 & 41.0 & 47.9 \\

\textsc{MTE}~\citep{fomicheva-etal-2020-unsupervised}
& 30.7 & 27.2 & 34.4 & 27.7 & 30.0
& 53.1 & 50.1 & 46.4 & 40.0 & 47.4 \\

\textsc{LexSim}~\citep{fomicheva-etal-2020-unsupervised}
& 28.7 & 29.8 & 33.1 & 26.8 & 29.6
& 54.9 & 51.5 & 46.1 & 38.1 & 47.6 \\

\textsc{DegMat}~\citep{lin2024generating}
& 30.3 & 30.1 & 33.2 & 27.0 & 30.1
& 53.3 & 50.7 & 45.9 & 39.2 & 47.3 \\

\textsc{SemanticEntropy}~\citep{kuhn2023semantic}
& 30.4 & 29.6 & 33.6 & 25.8 & 29.8
& 55.9 & 50.7 & 45.9 & 38.9 & 47.9 \\

\textsc{SAR}~\citep{duan-etal-2024-shifting}
& 30.2 & 29.2 & 35.4 & 26.8 & 30.4
& 55.2 & 50.1 & 47.1 & 40.1 & 48.1 \\

\textsc{SemanticDensity}~\citep{qiu2024semanticdensity}
& 27.9 & 28.2 & 32.8 & 29.5 & 29.6
& 50.9 & 49.1 & 48.7 & 53.5 & 50.6 \\

\textsc{CoCoA-MSP}~\citep{vashurin2025cocoa}
& 30.1 & 29.1 & 33.9 & 26.1 & 29.8
& 54.6 & 49.4 & 46.6 & 40.2 & 47.7 \\

\textsc{CoCoA-PPL}~\citep{vashurin2025cocoa}
& 30.0 & 29.0 & 34.0 & 27.8 & 30.2
& 53.2 & 50.1 & 46.8 & 40.2 & 47.6 \\

\textsc{CoCoA-MTE}~\citep{vashurin2025cocoa}
& 29.7 & 29.2 & 33.9 & 27.6 & 30.1
& 53.9 & 49.8 & 46.3 & 39.9 & 47.5 \\

\midrule
Random PR-AUC
& 29.3 & 26.5 & 33.5 & 29.2 & 29.6
& 49.7 & 49.8 & 50.0 & 49.8 & 49.8 \\
\bottomrule
\end{tabular}%
}
\caption{Full uncertainty-quantification baseline results across the four local decisions. Values are PR-AUC multiplied by 100.}
\label{tab:uq_full_results}
\end{table*}

\section{Prompts for the Mitigation Pipeline}
\label{app:mitigation-prompts}

This appendix provides the GPT-5 prompts used in our
attribution-guided intervention pipeline. The attribution tree first
selects a failure branch, after which GPT-5 returns only the
branch-specific support. The placeholder \texttt{\{question\}} is
replaced with the original user question.

\begin{tcolorbox}[
  colback=white,
  colframe=black,
  boxrule=0.5pt,
  arc=1mm,
  left=1mm,
  right=1mm,
  top=0.8mm,
  bottom=0.8mm,
  title=Unknown entity
]
\small
Identify the main entity in the image.

Return only:

Entity: \textless name or unknown\textgreater

Details: \textless 1-2 short sentences describing the entity\textgreater
\end{tcolorbox}

\begin{tcolorbox}[
  colback=white,
  colframe=black,
  boxrule=0.5pt,
  arc=1mm,
  left=1mm,
  right=1mm,
  top=0.8mm,
  bottom=0.8mm,
  title=Visual-evidence failure
]
\small
The image was degraded. Please make the image clearer, keeping main entities and fine-grained details if any.
\end{tcolorbox}

Before either factual branch, we first rerun the target VLM with an
entity-identification prompt. For PopVQA, we ask:
\textit{What is the entity in the photo? Mention the proper name of
the person/place/brand.} For iNaturalist, we ask:
\textit{What is the entity in the photo? Mention the common name of
the species.} The predicted entity is then passed to GPT-5 together
with the original question, without the image.

For \textsc{unknown fact}, GPT-5 provides short factual evidence
relevant to the question and the predicted entity.

\begin{tcolorbox}[
  colback=white,
  colframe=black,
  boxrule=0.5pt,
  arc=1mm,
  left=1mm,
  right=1mm,
  top=0.8mm,
  bottom=0.8mm,
  title=Unknown fact
]
\small
Provide factual evidence for answering the question using only the
entity name predicted by the target VLM.

Question: \texttt{\{question\}}

Predicted entity: \texttt{\{predicted\_entity\}}

Return only:

Facts: \textless 1-3 short facts relevant to the question and the
predicted entity\textgreater
\end{tcolorbox}

For \textsc{unrecallable fact}, GPT-5 rewrites the question so that
the predicted entity is stated explicitly. It is instructed not to
add factual information or answer the question.

\begin{tcolorbox}[
  colback=white,
  colframe=black,
  boxrule=0.5pt,
  arc=1mm,
  left=1mm,
  right=1mm,
  top=0.8mm,
  bottom=0.8mm,
  title=Unrecallable fact
]
\small
Rewrite the question so that the entity name predicted by the target
VLM is stated explicitly. Do not add factual information or answer
the question.

Question: \texttt{\{question\}}

Predicted entity: \texttt{\{predicted\_entity\}}

Return only:

New question: \textless rewritten question with the predicted entity
stated explicitly\textgreater
\end{tcolorbox}

\paragraph{Prompt used for rerunning the target VLM.}
For all non-visual branches, the target VLM is rerun with the original image and the following augmented prompt, where \texttt{\{manager\_text\}} is the GPT-5 output from the corresponding prompt above.

\begin{tcolorbox}[
  colback=white,
  colframe=black,
  boxrule=0.5pt,
  arc=1mm,
  left=1mm,
  right=1mm,
  top=0.8mm,
  bottom=0.8mm,
  title=Augmented target-VLM prompt
]
\small
\texttt{\{question\}}

Additional information:

\texttt{\{manager\_text\}}

Answer the original question using the image and the additional information.
\end{tcolorbox}

For \textsc{visual-evidence failure}, the prompt remains the original question; only the repaired image is used when rerunning the target VLM.

\section{Cross-Dataset Generalization}
\label{app:cross-dataset}

We study whether the failure-attribution probes transfer across dataset domains. Prior work has shown that VQA performance can degrade substantially under distribution shift, including shifts in answer priors~\citep{agrawal2018dont} and cross-dataset visual-language distributions~\citep{akula-etal-2021-crossvqa}. Given that PopVQA and iNaturalist differ in both entity domain and failure-mode distribution, we evaluate whether the attribution probes transfer between the two. We use the best feature family for each local decision (\S\ref{sec:53}) and train probes on PopVQA data only. \Cref{tab:cross_dataset_generalization} reports PopVQA-to-iNaturalist transfer. As expected, performance drops relative to the in-domain results, but the drop is not uniform across decisions. Visual-evidence attribution transfers best, reaching 84.9 average PR-AUC, consistent with the fact that this decision depends mainly on whether image degradation removes recognizable evidence. Recognition also remains above 70 average PR-AUC. In contrast, answer success and factual access are weaker, reaching 51.9 and 50.8 respectively. This suggests that image-side failure signals are more transferable across datasets, while factual decisions are more sensitive to the entity and knowledge distribution of the target domain.

\begin{table}[H]
\centering
\small
\setlength{\tabcolsep}{4pt}
\renewcommand{\arraystretch}{0.95}
\resizebox{\columnwidth}{!}{%
\begin{tabular}{llccccc}
\toprule
\textbf{Decision} & \textbf{Feature}
& \textbf{Gemma} & \textbf{Qwen2.5} & \textbf{Qwen3} & \textbf{Llama} & \textbf{Avg.} \\
\midrule
Recognition 
& \textsc{Vis}, Lin.
& 74.8 & 78.6 & 71.9 & 73.5 & 74.7 \\

Visual evidence 
& \textsc{Vis}, Lin.
& 86.7 & 88.9 & 79.8 & 84.1 & 84.9 \\

Answer success 
& \textsc{Last8}, Lin.
& 50.8 & 52.1 & 56.3 & 48.5 & 51.9 \\

Factual access 
& \textsc{Last8}, Tr.
& 51.2 & 49.7 & 53.1 & 49.0 & 50.8 \\
\bottomrule
\end{tabular}%
}
\caption{Cross-dataset generalization from PopVQA to iNaturalist. Probes are trained on PopVQA and evaluated directly on iNaturalist. Scores are PR-AUC \%.}
\label{tab:cross_dataset_generalization}
\end{table}

\section{Detailed Data Curation and Label Construction}
\label{app:data-curation-details}

This appendix provides the complete procedure used to curate the
evaluation pairs and assign model-specific operational outcomes. All
tests are performed separately for each target VLM. Consequently, the
same image--question pair may receive different labels for different
models.

\subsection{Source Datasets and Initial Sampling}
\label{app:data-sources}

We construct the dataset from PopVQA
\citep{cohen-etal-2025-performance} and the iNaturalist subset of
Encyclopedic VQA
\citep{vanhorn2018inaturalist,mensink23iccv}. PopVQA contains factual
questions associated with four broad entity categories: celebrities,
landmarks, logos, and paintings. The iNaturalist subset contains
fine-grained factual questions about plant and animal species.

We initially sample $6{,}300$ image--question pairs from PopVQA and
$4{,}400$ pairs from iNaturalist. We retain the original reference
answers and entity annotations supplied by the datasets. Examples of
the resulting entity-linked factual questions are shown in
Appendix~\ref{app:sample-qa}.

\subsection{Question-Type Filtering}
\label{app:question-filtering}

We remove question forms that can often be answered through coarse
linguistic priors or elimination without requiring reliable entity
recognition. Specifically, we exclude:

\begin{itemize}
    \item binary questions beginning with \emph{Does}, \emph{Can}, or
    \emph{Is}; and
    \item disjunctive questions containing an explicit \emph{or}.
\end{itemize}

The remaining questions require open-ended factual answers and are
therefore better suited to separating recognition from factual access.

\subsection{Image-Ablation Filtering}
\label{app:image-ablation}

We next remove questions that do not meaningfully depend on the image.
For each image--question pair, we replace the original image with a
uniform white canvas while keeping the question unchanged. Each target
VLM is then evaluated using the same answer-scoring protocol as for the
original input. If any target model answers the question correctly under image
ablation, we treat the item as insufficiently image-grounded and remove
it from the shared evaluation set. When the underlying evaluation contains alternative answer orders, we apply the ablation across the same answer permutations used in the original evaluation and remove the item if the model succeeds under the ablated input. 

After question-type and image-ablation filtering, the retained dataset
contains $4{,}863$ PopVQA pairs and $2{,}116$ iNaturalist pairs.

\subsection{Entity-Recognition Test}
\label{app:recognition-protocol}

Open-ended entity naming can be difficult to evaluate consistently
because a response may contain aliases, partial names, broader
categories, or descriptive paraphrases. We therefore use a controlled
yes/no verification protocol.

For each image, we construct one positive recognition prompt containing the ground-truth entity name and three negative prompts containing distractor entities. The positive prompt uses the following generic template:

\begin{quote}
\ex{Is the [entity type] in the image [ground-truth entity]?}
\end{quote}

The entity-type wording is adapted to the corresponding dataset
category, for example, person, place, logo, painting, plant, insect, or
animal.

Distractors are sampled from the same semantic subtype as the
ground-truth entity. For iNaturalist, we sample alternatives from the
same broad biological group, such as replacing one mammal with another
mammal or one insect with another insect. For PopVQA, distractors are
sampled from the same entity category, such as celebrity, landmark,
logo, or painting. This prevents the negative prompts from being
trivially rejected based on category mismatch.

An entity is considered reliably recognized only when the target VLM:

\begin{itemize}
    \item answers \textit{yes} to the prompt containing the
    ground-truth entity; and
    \item answers \textit{no} to all three distractor prompts.
\end{itemize}

If either condition is violated, the example receives the operational
label \textsc{unknown entity}. This label indicates failure under the
verification protocol. It does not establish that the entity is
completely absent from the model's internal representations or
pretraining knowledge.

\subsection{Controlled Visual-Evidence Construction}
\label{app:visual-evidence-construction}

PopVQA and iNaturalist primarily contain clear, curated images and do
not provide sufficient naturally occurring examples of poor visual
evidence. We therefore construct a controlled recognizability stress
test from examples that initially pass the entity-recognition
protocol.

For each reliably recognized image, we progressively apply a
combination of:

\begin{itemize}
    \item Gaussian blur;
    \item additive Gaussian noise;
    \item JPEG compression; and
    \item spatial downsampling followed by resizing to the original
    model input resolution.
\end{itemize}

Corruption severity increases monotonically across predefined levels. \Cref{tab:corrupt} provides full configurations for the adaptive corruption schedule used to construct recognition-boundary examples. Each level applies a fixed bundle of perturbations: Gaussian blur, additive Gaussian noise, JPEG compression, and downsampling (followed by resizing back to the model input size).

\begin{table}[H]
\centering
\small
\setlength{\tabcolsep}{5pt}
\begin{tabular}{lcccc}
\toprule
\textbf{Level} & \textbf{Blur} & \textbf{Noise (std)} & \textbf{JPEG} & \textbf{Downscale} \\
\midrule
1 & 2.5 & 25 & 60 & 0.75 \\
2 & 3.5 & 35 & 45 & 0.60 \\
3 & 4.5 & 45 & 35 & 0.50 \\
4 & 5.5 & 60 & 25 & 0.40 \\
\bottomrule
\end{tabular}
\caption{Corruption levels used in the adaptive schedule. ``Blur'' is the Gaussian blur $\sigma$, ``Noise'' is the standard deviation of additive Gaussian noise in pixel space, ``JPEG'' is the compression quality factor, and ``Downscale'' is the downsampling ratio prior to resizing back to the model input resolution.}
\label{tab:corrupt}
\end{table}

After applying each level, we repeat the positive entity-recognition prompt using the ground-truth entity name. We retain the first corrupted version for which the model changes from successful recognition to recognition failure. This example receives the label \textsc{visual-evidence failure}. Formally, if the original image passes the recognition test and corruption level $j$ is the first level at which it fails, the image produced at level $j$ becomes the corresponding visual-evidence example. Images that remain recognizable at the maximum corruption level are not assigned to this category. This procedure ensures that visual-evidence examples are derived from images that the same target model could originally recognize, rather than from arbitrary noisy images.

Because the number of recognition-boundary examples can exceed the number of naturally occurring \textsc{unknown entity} examples, we downsample the former so that the two recognition-failure classes have equal size. This balancing is performed independently for each target VLM.

The resulting category should be interpreted narrowly. It represents recognition failure induced by controlled degradation and does not
cover all natural visual-evidence problems, including occlusion, unusual viewpoints, low lighting, cropping, poor framing, or domain-specific imaging artifacts.

\subsection{Answer-Success Evaluation}
\label{app:answer-success-labeling}

For every example that passes the entity-recognition test, we evaluate
the target VLM on the original knowledge-intensive question using the
original image. If the generated answer is judged correct, the example
receives the label \textsc{success} and no additional attribution test
is required.

Because generated responses may contain full sentences, aliases,
minor spelling variations, or equivalent descriptions, we use a
text-only LLM judge when exact matching is insufficient. The judge
receives the original question, the set of acceptable reference
answers, and the candidate response. It is instructed to evaluate only
the core factual content and to accept minor lexical variants when
their meanings match the reference. The complete judge prompt is
provided in Appendix~\ref{app:eval_judge}.

\subsection{Entity-Explicit Factual-Access Test}
\label{app:factual-access-protocol}

For examples in which the entity is reliably recognized but the
original factual question is answered incorrectly, we test whether the
failure is resolved when the entity name is explicitly included in the
question. This intervention follows
\citet{cohen-etal-2025-performance}. For example, an original question such as:

\begin{quote}
\ex{What year was this actor born?}
\end{quote}

is rewritten as:

\begin{quote}
\ex{What year was Tom Cruise born?}
\end{quote}

The original image remains available to the target VLM, although the
rewritten question no longer requires the model to infer the entity
name from the image. Apart from replacing the referential expression
with the annotated entity name, we preserve the factual content of the
question.

The result of this intervention determines the final operational
label:

\begin{compactitem}
    \item If the model fails on the original question but succeeds on
    the entity-explicit version, the example receives
    \textsc{unrecallable fact}.
    \item If the model fails on both the original and entity-explicit
    versions, the example receives \textsc{unknown fact}.
\end{compactitem}

The term \textsc{unrecallable fact} denotes a failure that is resolved
when the entity is made explicit. The term \textsc{unknown fact}
denotes a failure that persists under this intervention. These labels
do not prove whether a fact is encoded in the model's parameters; they
summarize behavior under the controlled entity-explicit test.

\subsection{Model-Specific Attribution Datasets}
\label{app:model-specific-datasets}

The complete assignment procedure is applied independently to each
target VLM. For a given model, each retained image--question pair is
assigned one of the following outcomes:

\begin{align}
Y \in \{
&
\textsc{visual-evidence failure}, \nonumber\\
&
\textsc{unknown entity}, \nonumber\\
&
\textsc{success}, \nonumber\\
&
\textsc{unknown fact}, \nonumber\\
&
\textsc{unrecallable fact}
\}.
\end{align}

The assignment follows the attribution tree:

\begin{itemize}
    \item failing the entity-recognition test yields
    \textsc{unknown entity};
    \item losing recognition under controlled degradation yields
    \textsc{visual-evidence failure};
    \item passing recognition and answering the original question
    correctly yields \textsc{success};
    \item failing the original question but succeeding after the
    entity-explicit intervention yields \textsc{unrecallable fact};
    \item failing both factual-question variants yields
    \textsc{unknown fact}.
\end{itemize}

Because recognition and factual behavior differ across target VLMs,
the resulting attribution datasets are model-specific. These datasets
are subsequently divided into stratified training, validation, and test
splits for fitting the four local attribution probes.

\end{document}